\RequirePackage{amsmath}
\RequirePackage{fix-cm}
\documentclass[smallextended]{svjour3}       
\smartqed  
\usepackage{graphicx}
\usepackage{realboxes}
\usepackage{csvsimple}
\usepackage{datatool}
\usepackage{geometry}
\usepackage{booktabs}
\usepackage{varwidth}
\usepackage{array}
\usepackage[table]{xcolor}
\usepackage{tabularx}
\usepackage[ruled,vlined, algosection, procnumbered]{algorithm2e}
\usepackage{placeins}
\usepackage{subcaption}
\usepackage[utf8]{inputenc}

\arrayrulecolor{black!30}
\newcolumntype{?}{!{\color[gray]{0.7}\vrule width .4pt}}
\newcolumntype{P}[1]{>{\centering\arraybackslash}p{#1}}
\usepackage[pagewise]{lineno}
\usepackage[backend=biber, citestyle=numeric, bibstyle=authoryear, bibencoding=utf8, maxbibnames=100]{biblatex}
\usepackage{rotating}

\makeatletter
\input{numeric.bbx}
\makeatother

\begin{document}
	
	\title{An Eager Splitting Strategy for Online Decision Trees in Ensembles
	}
	
	
	\author{Chaitanya Manapragada \and
		Heitor M Gomes \and 
		Mahsa Salehi \and
		Albert Bifet \and 
		Geoffrey I Webb       
	}
	
	
	\institute{Chaitanya Manapragada, Geoff Webb, Mahsa Salehi \at
		Monash University, Australia \\
		\email{FirstName.LastName@monash.edu}           
		\and
		{Albert Bifet, Heitor Murilo Gomes} \at
		University of Waikato, New Zealand \\
		\email{FirstName.LastName@waikato.ac.nz}              
	}

	\date{Received: date / Accepted: date}

	\maketitle

	\begin{abstract}
		
		
		Decision tree ensembles are widely used in practice. In this work, we study in ensemble settings the effectiveness of replacing the split strategy for the state-of-the-art online tree learner, Hoeffding Tree, with a rigorous but more eager splitting strategy that we had previously published as Hoeffding AnyTime Tree. Hoeffding AnyTime Tree (HATT), uses the Hoeffding Test to determine whether the current best candidate split is superior to the current split, with the possibility of revision, while Hoeffding Tree aims to determine whether the top candidate is better than the second best and if a test is selected, fixes it for all posterity. HATT converges to the ideal batch tree while Hoeffding Tree does not. We find that HATT is an efficacious base learner for online bagging and online boosting ensembles. On UCI and synthetic streams,  HATT as a base learner outperforms HT within a 0.05 significance level for the majority of tested ensembles on what we believe is the largest and most comprehensive set of testbenches in the online learning literature. Our results indicate that HATT is a superior alternative to Hoeffding Tree in a large number of ensemble settings.
		
		
	\end{abstract}
	
	\section{Introduction}
	\label{intro}

	Hoeffding Tree \cite{domingos2000mining} is the base learner for online versions of widely successful ensembling methods such as random forests, bagging and boosting. A host of derivative online ensemble methods have been proposed in the literature; the methods vary in the diversity of base learners they use, the adaptability of the ensemble as a whole and of the component base learners. It is imperative that a proposed method works well in ensemble settings as well as standalone, as likely usage will often involve ensemble learners. 
	
	Hoeffding AnyTime Tree (HATT) \cite{manapragada2018extremely} is based on a simple and fundamental change to Hoeffding Tree; HATT uses the current best available split attribute at a node until a better one is found, as opposed to Hoeffding Tree which aims to find a split attribute that will never have to be replaced. HATT converges to the ideal batch tree while Hoeffding Tree does not. The Extremely Fast Decision Tree (EFDT) implementation of HATT has demonstrated significant improvement in prequential accuracy\footnote{In the prequential setting, training instances arrive in a sequence, and the true target value pertaining to each training instance is made available after the predictor has offered a prediction for a sequence of $n$ instances. The loss function applied is necessarily incremental in nature. Choosing $n=1$ --- that is, evaluating and then updating the predictor after every instance---is an obvious transformation of a periodic evaluation process into an instantaneous one. While not typical of real world application scenarios, prequential accuracy serves as a useful approximation thereto.} on UCI data streams over the Hoeffding Tree implementation Very Fast Decision Tree (VFDT) from the MOA toolkit \cite{bifet2010moa} in a comparison setting that included only a minor change to VFDT in order to obtain EFDT. 
	
	The current state-of-art tree-based online classification learners are ensembles that use VFDT as a base learner. This work provides a detailed assessment of the relative performance of VFDT and EFDT as base learners for online ensembling techniques.

	The ensemble methods we test generally far outperform plain EFDT and VFDT in terms of prequential accuracy, which is to be expected; bagging approaches specifically utilize diversity in order to aim to ``stabilize'' the predictive model, while boosting approaches modify the model in a principled manner to focus on learning misclassified examples. Generally speaking, the success of ensembling strategies on this testbench and in practice in the real world demonstrates that tree learners may suffer from high variance and a small considered hypothesis space (bias) when used as individual classifiers. Further, it should be noted that a formal definition of overfitting or underfitting is difficult to obtain in an online setting that may include concept drift unless the nature and magnitude of drift is bounded and fixed in a highly restrictive manner.
	
	It is the prerogative of the user to pick an ensemble method that suits the application of interest; a corollary of the No Free Lunch Theorems is that no individual system will obtain superior performance on all possible problems \cite{wolpert1997no, wolpert2005coevolutionary, wolpert96}. With this proviso, in this work, we show that on the most common testbenches, it is highly beneficial in terms of prequential accuracy to use HATT as a base classifier for ensemble methods in the place of Hoeffding Tree.
	
	The main contributions of this paper are:
	\begin{enumerate}
		\item A comprehensive experimental analysis comparing EFDT and VFDT as base learners for a large set of online ensembles across a diverse set of real and synthetic streams, building on our preliminary work in \cite{manapragada2018extremely}
		\item Identification of scenarios in which EFDT has a definitive advantage with high confidence
		\item A working hypothesis for the observed outperformance of EFDT (Section \ref{sec:hattensemble})
		\item Discussion of how change detection mechanisms may interact adversely with ensemble components
	\end{enumerate}

	This paper is organized as follows: Section \ref{background} presents a general overview of the context of stream learning within which this work is placed. Section \ref{sec:baselearners} presents HATT. This is largely similar to our previous presentation in \cite{manapragada2018extremely}, but with the addition of a new subsection on HATT in  ensembles (Section \ref{sec:hattensemble}), the primary focus of this paper. Section \ref{expsetup} describes our experimental setup. Section \ref{expresults} is a discussion of experimental results in terms of the types of ensembles and settings in which HATT is advantageous as a base learner. Section \ref{conclusions} summarizes our findings.

	\begin{figure}
		\centering
		
		\begin{subfigure}[r]{1.0\textwidth}
			\includegraphics[trim={0 10cm 15cm 2cm},clip,width=\textwidth]{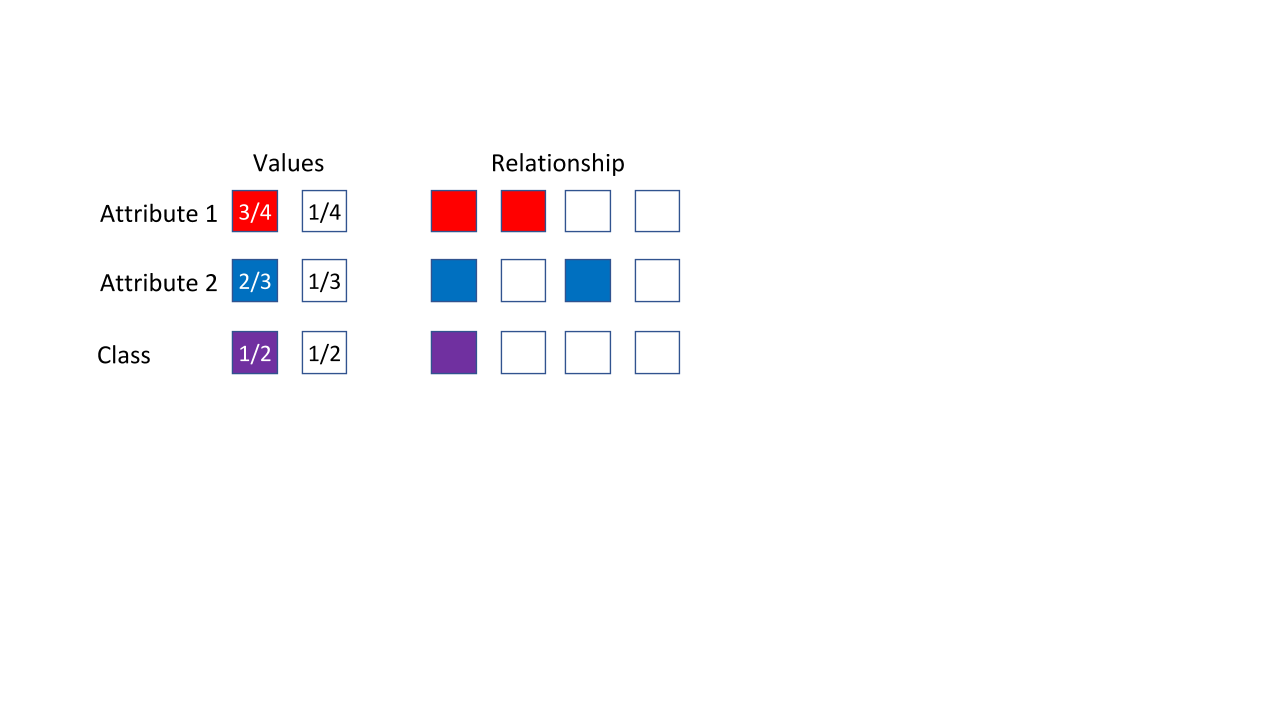}
			\caption{Data Types --- Data with Attribute 1 = Red and Attribute 2 = Blue are classified as Purple. All other data are classified as White. Attributes are independent. \\
				Distribution of Data --- Attribute 1 is red 3/4 of the time, Attribute 2 is blue 2/3 of the time, so each class occurs 1/2 the time. }		\label{fig:frequency}
		\end{subfigure}
		
		\begin{subfigure}[b]{1.0\textwidth}
			\includegraphics[width=\textwidth]{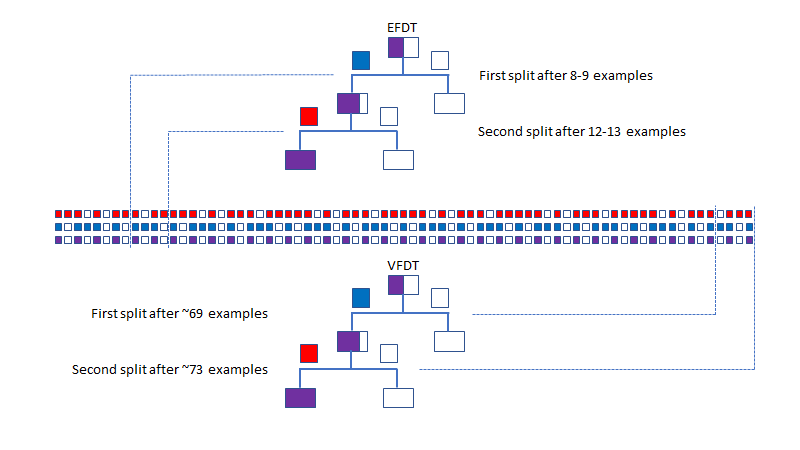}
			\caption{Streaming Example with Tree Construction. }
			\label{fig:stream}
		\end{subfigure}
		\caption{On a randomized stream generated from the data schema and distribution shown above in (a), EFDT first splits after 8-9 examples, then splits again after 12-13 examples to build the correct tree. VFDT takes around 69 examples for the first split, and makes the second split at around 73 examples. EFDT greatly increases statistical efficiency without compromising the use of a rigorous statistical test to determine split attributes, and revises splits in order to converge to the ideal batch tree.}\label{fig:efdtexample}
	\end{figure}

	\section{Background}
	\label{background}
	
	The traditional batch learning setting for machine learning is designed for finite datasets drawn from stationary distributions. Methods developed for learning from such datasets do not readily lend themselves to modern data processing applications dealing with streams of data where instances arrive continuously, generated by processes that may themselves be ever-changing. It is necessary to design new algorithms for learning from such settings, and a good place to start is from algorithms designed for batch settings. Decision Trees have been ubiquitous in batch learning settings, both as individual learners and in multiple ensembled forms such as Random Forests.

	\subsection{Decision Trees for Batch Learning}

	Of the many approaches to inductive learning, Decision Trees are a particularly utile paradigm that store knowledge in an easily interpretable manner. Algorithms that build decision tree models recursively divide the sample space with hyperplane decision boundaries. Each division represents a conditioning of the sample space on a particular set of data attribute values or ranges. The knowledge obtained from a Decision Tree is represented in an elementary form; in the classification case, each path down the tree results in a conditional probability distribution $P(C | X_1=v_1, X_2=v_2, ...)$, that is, the probability distribution of the class values given the observations $X_1=v_1$, $X_2=v_2, ...$. The regression case may use, for instance, a simple average of observed target values.
	
	Decision Trees are a natural starting point for the study of extending inductive strategies to streaming scenarios; their simplicity allows us to compute model complexity \cite{de2019measuring}, and they are highly interpretable. Concept Learning System (CLS) \cite{earl1966experiments} was one of the first decision tree algorithms, published in 1966. The next decision tree system of note was Iterative Dichotomizer (ID3), published in 1979 \cite{quinlan1979discovering, quinlan1983learning}. ID3 introduced the idea of using information gain as a split heuristic, though it was limited to handling binary classification and assumed all instances were correctly labelled.
	
	``Classification and Regression Trees'' (CART), proposed in 1984, included pruning to adjust for overfitting, multiclass classification, and a solution for regression \cite{breiman1984classification}. CART used the Gini coefficient as a heuristic, not Information Gain. Consequently, a major improvement over ID3 called C4.5 that also addressed pruning, multiclass classification and regression was released \cite{Quinlan:2031749}. The ideas embedded in C4.5 and CART form the basis of most modern decision tree learning systems today designed for both batch and streaming scenarios.

	\subsection{Incremental Decision Trees}

	Attempts at producing online versions of decision trees largely dominate work in online learning.
	
	Work on incremental decision trees began appearing just as batch decision trees started to mature. ID4 \cite{schlimmer1986case} was an incremental extension to ID3 that stored instances used for a split at each level of the tree. ID4 was conceived for a binary classification problem; the test of choice is the $\chi^2$ test to determine whether the attribute with maximal separation power is independent of the target variable. When confidence of dependence is reached with the $\chi^2$ test, the maximal attribute is split upon. The limitation of ID4 was that storage required was in the order of the number of instances.
	
	Storage is a primary issue to address in the construction of an incremental supervised learning, because streams may be assumed to be indefinite. Another key problem is determining when and how the algorithm should modify the model---unlike with batch learning, one does not have all training instances available in one go and must periodically decide whether one has enough data to modify the decision tree model with some degree of confidence.
	
	\sloppy Strategies that process each instance and discard it immediately afterwards---\textit{one-pass} strategies---would hypothetically address the problem of storage by only requiring storage of the order of the size of the tree, not the number of instances.

	\begin{algorithm}

		\DontPrintSemicolon
		\SetAlgoLined
		\KwIn{$S$, a sequence of examples,
			\\ \Indp\Indp $\mathbf{X}$, a set of discrete attributes,
			\\ G(.), a split evaluation function
			\\ $\delta$, one minus the desired probability of choosing the correct attribute at any given node
		}
		\KwOut{$HT$, a decision tree.}
		\Begin{
			Let HT be a tree with a single leaf $l_1$ (the $root$).\;
			Let $\mathbf{X_1} = \mathbf{X} \cup X_{\emptyset}$.\;
			Let $\overline{G_1}(X_{\emptyset})$ be the $\overline{G}$ obtained by predicting the most
			frequent class in $S$\;
			\ForEach {class $y_k$}{
				\ForEach { value $x_{ij}$ of each attribute $X_i \in \mathbf{X}$}{
					Let $n_{ijk}(l_1) = 0$\;
				}
			}
			
			\ForEach {example $(\vec{x},y)$ in S}{
				Sort $(\vec{x},y)$ into a leaf $l$ using $HT$\;
				\ForEach {$x_{ij}$ in $\vec{x}$ such that $X_i \in \mathbf{X}_l$}{
					Increment $n_{ijk}(l)$\;
				}
				Label $l$ with the majority class among the examples seen so far at $l$\;
				\If {the examples seen so far at $l$ are not all of the same class}{
					Compute $\overline{G_l}(X_i)$ for each attribute $X_i \in \mathbf{X}_l - \{X_{\emptyset}\}$ using the counts $n_{ijk}(l)$\;
					Let $X_a$ be the attribute with highest  $\overline{G_l}$\;
					Let $X_b$ be the attribute with second-highest  $\overline{G_l}$\;
					Compute $\epsilon$ using:
					\begin{equation} \label{eq:1}
					\epsilon = \sqrt[]{\frac{R^2 \log(1/\delta)}{2n}}
					\end{equation}					
					\If {$\overline{G_l}(X_a) - \overline{G_l}(X_b) > \epsilon$ and $X_a \neq X_{\emptyset}$}{
						Replace $l$ by an internal node that splits on $X_a$
						\ForEach {branch of the split}{
							Add a new leaf $l_m$ and let $\mathbf{X}_m = \mathbf{X} - \{X_\emptyset\}$
							Let $\overline{G}_m(X_\emptyset)$ be the $\overline{G}$ obtained by predicting the most frequent class at $l_m$
							\ForEach {class $y_k$ and each value $X_{ij}$ of each attribute $X_i \in \mathbf{X}_m - \{X_\emptyset\}]\}$}{
								Let $n_{ijk}(l_m) = 0$	
							}
						}
					}
				}			
			}
			Return HT
		}

		\caption{Hoeffding Tree, Domingos \& Hulten (2000)
			\Indp\Indp\Indp\Indp  \small{--Reproduced verbatim from original-- }
		} 
		\label{table:vfdt}
	\end{algorithm}
	
	In the batch learning scenario, all target values are available at the outset, barring missing values.	In the online setting, there is expected to be an infinite stream of instances and thus storage is considered impossible in the limit. Incremental tree learners are thus typically one-pass learners, in that they process each training instance exactly once. And because they are one-pass learners, incremental learners naturally allow for continuous evaluation; continuous evaluation of learning enables us, among other things, to detect or otherwise respond to concept drift in streams and adapt the learner accordingly.
	
	HoeffdingTree (Algorithm \ref{table:vfdt}) was one of several attempts \cite{schlimmer1986case, utgoff1989incremental} to provide a one-pass solution. Its success may be attributed to the fact that it was the first one-pass learner to also offer a robust solution to the problem of how the algorithm should modify the tree model. This robustness lies in the guarantees provided by the HoeffdingTree algorithm on the deviation of the inducted tree from the ideal batch tree---the hypothetical tree that would be learned if all infinite examples from a stationary distribution were made available at once. Hoeffding Tree uses a statistical test---the Hoeffding Test \cite{domingos2000mining, hoeffding1963probability}---to determine the most appropriate time to split. Its success may be attributed to the fact that it provided both a one-pass solution and deviation guarantees in the same package.
	
	The ideas that underlie HoeffdingTree were individually and independently developed in related contexts. Work on scalability of batch learners had helped set the foundation for one-pass learning in sequential prediction scenarios. Bootstrapped Optimistic Algorithm for Tree construction (BOAT) \cite{Gehrke:1999:BDT:304182.304197} in particular lays the groundwork of ideas for a tree refined in stages, though it deals with a batch setting. BOAT represents a typical attempt at learning from a large database that does not use a predictive sequential setting, by sampling fixed size chunks that are used to bootstrap multiple trees. A ``coarse" tree is then extracted, based on the overlapping parts of the bootstrapped trees in terms of split decisions; this tree is further refined to produce a final tree by passing the whole dataset over it. The system is ``incremental" in the sense that it can process additional datasets; and it is responsive to drift in that the system detects when a new dataset requires a change in split criterion at a node through a global assessment of split criterion, and causes a rebuild of the subtree rooted at that node. While key ideas that shape later trees are developed in this work, the sizes of the initial bootstrap samples are arbitrarily chosen, and concurrently the notion of anytime prediction is not entertained---there is no automated way of determining how many examples suffice to build a first reliable tree. Further, the focus is on minimizing utilization of main memory; it is assumed that the database $D$ is available for a corrective step in the algorithm. On the other hand, Hoeffding Tree is truly one-pass, in that it is assumed that an example is seen only once, then discarded. Meanwhile, the RainForest framework \cite{gehrke2000rainforest} introduces the idea of storing attribute-value-class counts at nodes, which we see in Hoeffding Tree as \textit{node statistics}. Node statistics are indispensable in HoeffdingTree; they solve the one-pass problem and are critical in the application of the Hoeffding Test for split evaluation.
	
	Hoeffding Tree may be considered to be aiming to achieve the same evaluation objective as CART and C4.5---maximal class purity---with a heuristic designed to be relevant in the streaming scenario where the data generation distribution is assumed to be static.

	Hoeffding Tree has dominated the development of incremental learning, winning a KDD Test of Time award \cite{kddtestoftimedomingos} in 2015. The statistical test it uses---the Hoeffding Test---is known to possibly lead to loose bounds, that is, the bound on the difference between the observed mean and the true mean is larger than would have been obtained with a tighter bound on the standard deviation. $R$, the range (or support) of the random variable may be considered too large an upper bound for the standard deviation in cases where variance is small \cite{looseHoeffding}. The population of random variables\footnote{There is a common misconception that an individual random variable ``changes, taking on a number of values during a process''; in fact, a process is a sequence of events, each of which corresponds to an individual random variable that has taken a particular value (which is fixed and never to change).} consists of the difference of cumulatively averaged information gain measurements of the top two split attributes after each learning step; a loose bound implies that a far larger number of such measurements is needed than actually required to establish the winning attribute \cite{heidrich2009hoeffding}. Because the bound depends on the size of the population of random variables, which in turn depends directly on the number of observations of training instances, greater statistical efficiency (see Figure \ref{fig:efdtexample}) may be achieved by either using an alternative test or by changing the application of the test. Our strategy, Hoeffding AnyTime Tree (HATT) \cite{manapragada2018extremely}, is an example of the latter.
	
	To the best of our knowledge, Hoeffding Tree is the first attempt at incremental tree learning that provides guarantees of bounded divergence from a theoretical batch tree that has all examples in an infinite stream available to it at once. These guarantees are useful in that they enable us to compare theoretical performance with respect to longstanding, reliable decision tree methods. However, such guarantees often assume each data instance corresponds to a random variable from a stationary process that generates independent and identically distributed ($i.i.d$) random variables. This is usually not a valid assumption in streaming scenarios in which processes can change over time, and working around this assumption would require placing restrictions on the nature of change of the generating process. Such restrictions may be meaningful in the context of particular, well-studied processes where the nature of change is fully known.

	\subsection{Ensembles}
	\label{sec:ensembles}
	
	Ensemble methods were proposed as a means of utilizing multiple predictors to represent and combine various parts of the instance space, for various heuristic reasons---using a set of base predictors may reduce the risk of building an overfitted individual predictor; reducing the complexity of each individual base predictor may also serve to reduce risk from overfitting \cite{dietterich2000ensemble}. Further, using multiple base predictors may allow us to achieve a rather different bias-variance profile compared to using a single predictor.
	
	Bootstrap aggregation- or ``bagging'' ensembles were proposed for stability, that is, to limit variance \cite{breiman1996bagging}. By ``perturbing'' the learning set for each component of an ensemble of predictors, Breiman creates predictors that effectively represent a larger bias (the set of hypothesis functions being considered) through combination. Assuming then a stationary distribution over the instance space, the ensemble is likely to demonstrate lower dataset dependent variance than a single tree. The objective for bagging differs from boosting in its addressal of controlling variance, and the accompanying heuristic follows.
	
	Boosting was first proposed as an approach to answer affirmatively the question of whether a combination of weak predictors---predictors that find hypotheses that perform only slightly better than random guessing---can learn strongly, that is, are able to find hypotheses ``that are correct with high probability on all but an arbitrarily small fraction of instances'' \cite{schapire1990strength}. In the classification setting, boosting works by using some components of the ensemble to focus on weighting misclassified examples higher so they are preferentially learned with respect to their actual frequency of occurrence, and thus potentially classified better on a test set if they were not merely noisy instances. In practice, a major advantage is that ensemble components need to be less complex (thus less prone to individually overfit). The objective is to reduce the possibility of overfitting, and the heuristic is the idea that using individual models of lower complexity will mitigate the danger of overfitting while remaining effective as constituent ensemble components would have learned subconcepts that would otherwise have been learned by a single complex model.

	Online versions of both boosting (OzaBoost) and bagging (OzaBag) were proposed in \cite{oza05}. Hoeffding Tree versions of the ensembles were made available in MOA \cite{bifet2010moa}.
	
	Because the online versions assume infinite streams, strategies to sample the input space in a manner equivalent to sampling from a finite sample space had to be devised. Online bagging as performed by OzaBag achieves the goal of providing each learner with a different subset of the sample space by weighting each example with a value drawn from a Poisson(1) distribution. Similarly, online boosting as performed by OzaBoost provides the first base predictor an example weighted as $Pois(\lambda=1)$; the second base predictor receives weighted examples with $\lambda$ adjusted so that misclassified examples comprise half the total weight, thus heavily stressing the learning of misclassified examples. This telescoping sequence of boosting misclassified examples is continued through the predictors in the ensemble.
	
	While ensembles do tend to be more expensive in terms of memory and time than individual predictors, their ability to provide diversity  particularly motivates their usage with evolving streams. As discussed, predictor diversity is central to bagging approaches; it plays a role in boosting approaches; in evolving scenarios, some base predictors may learn the latest versions of a changing concept, some may preserve portions of a concept that may recur, and some predictors could be reset if they have not been of utility for a prolonged period. Note that there also exists the possibility that an ensemble approach may degenerate its base predictors into a set of redundant models \cite{gomes2017survey}.

	\section{Hoeffding AnyTime Tree (HATT)}
	\label{sec:baselearners}

	\begin{algorithm}
		\DontPrintSemicolon
		\SetAlgoLined
		\KwIn{$S$, a sequence of examples. At time t, the observed sequence is $S^t = ((\vec{x}_1, y_1), (\vec{x}_2, y_2), ... (\vec{x}_t, y_t))$ 
			\\ \Indp\Indp $\mathbf{X} = \{X_1, X_2... X_m\}$, a set of $m$ attributes
			\\  $\delta$, the acceptable probability of choosing the wrong split attribute at a given node
			\\ G(.), a split evaluation function
		}
		\KwResult{$HATT^t$, the model at time $t$ constructed from having observed sequence $S^t$.}
		\Begin{
			Let HATT be a tree with a single leaf, the $root$ \;
			Let $\mathbf{X_1} = \mathbf{X} \cup {X_\emptyset}$\;
			Let $G_1(X_{\emptyset})$ be the $G$ obtained by predicting the most
			frequent class in S\;
			\ForEach {class $y_k$}{
				\ForEach { value $x_{ij}$ of each attribute $X_i \in \mathbf{X}$}{
					Set counter $n_{ijk}(root) = 0$\;
				}
			}
			
			\ForEach {example $(\vec{x},y)$ in S}{
				Sort $(\vec{x},y)$ into a leaf $l$ using $HATT$\;
				\ForEach {node in path $(root ... l)$}{
					\ForEach {$x_{ij}$ in $\vec{x}$ such that $X_i \in \mathbf{X}_{node}$}{
						Increment $n_{ijk}(node)$\;
						\eIf{$node = l$}{
							$AttemptToSplit(l)$
						}
						{
							$ReEvaluateBestSplit(node)$
						}
						
					}
				}
			}
		}
		\caption{Hoeffding Anytime Tree \label{table:HATT}}
	\end{algorithm}

	\begin{function}
		\DontPrintSemicolon
		\SetAlgoLined
		\Begin{
			Label $l$ with the majority class at $l$\;
			\If {all examples at $l$ are not of the same class}{
				Compute $\overline{G_l}(X_i)$ for each attribute $\mathbf{X}_l - \{\mathbf{X_{\emptyset}}\}$ using the counts $n_{ijk}(l)$\;
				Let $X_a$ be the attribute with the highest $\overline{G_l}$\;
				Let $X_b = X_{\emptyset}$\;	
				Compute $\epsilon$ using equation \ref{eq:1}\;
				\If {$\overline{G_l}(X_a) - \overline{G_l}(X_b) > \epsilon$ and $X_a \neq X_{\emptyset}$}{
					Replace $l$ by an internal node that splits on $X_a$\;
					\For {each branch of the split}{
						Add a new leaf $l_m$ and let $\mathbf{X}_m = \mathbf{X} - X_a$\;
						Let $\overline{G_m}(X_{\emptyset})$ be the $G$ obtained by predicting the most frequent class at $l_m$\;
						\For {each class $y_k$ and each value $x_{ij}$ of each attribute $X_i \in X_m -$  $\{X_{\emptyset}\}$}{
							Let $n_{ijk}(l_m) = 0$.
						} 
					}		
				}
			}
		}
		\caption{AttemptToSplit(leafNode $l$)\label{table:attempttosplit}}
	\end{function}
	
	\begin{function}
		\DontPrintSemicolon
		\SetAlgoLined
		\Begin{
			Compute $\overline{G}_{int}(X_i)$ for each attribute $\mathbf{X}_{int} - \{\mathbf{X_{\emptyset}}\}$ using the counts $n_{ijk}({int})$\;
			Let $X_a$ be the attribute with the highest $\overline{G}_{int}$\;
			Let $X_{current}$ be the \textit{current} split attribute\;	
			Compute $\epsilon$ using equation \ref{eq:1}\;
			\If {$\overline{G}_l(X_a) - \overline{G}_l(X_{current}) > \epsilon$}{
				\uIf{$X_a = X_{\emptyset}$}{
					Replace internal node $int$ with a leaf (kills subtree)\;
				}
				\ElseIf{$X_a \neq X_{current}$}{
					Replace $int$ with an internal node that splits on $X_a$\;
					\For {each branch of the split}{
						Add a new leaf $l_m$ and let $\mathbf{X}_m = \mathbf{X} - X_a$\;
						Let $\overline{G}_m(X_{\emptyset})$ be the $G$ obtained by predicting the most frequent class at $l_m$\;
						\For {each class $y_k$ and each value $x_{ij}$ of each attribute $X_i \in X_m -$  $\{X_{\emptyset}\}$}{
							Let $n_{ijk}(l_m) = 0$.
						}
					}	
				}
			}

		}	
		\caption{ReEvaluateBestSplit(internalNode $int$)\label{table:reevalsplit}}
	\end{function}

	Hoeffding AnyTime Tree (Algorithm \ref{table:HATT}) \cite{manapragada2018extremely} makes a simple change to the attribute selection mechanism of Hoeffding Tree and achieves better performance on a large testbench. Given the streaming setting, the objective underwent a subtle change; convergence to the ideal batch tree was the aim, as divergence from the ideal batch tree grows as the number of leaves in HoeffdingTree, which may plausibly be expected to see exponential growth, invalidating it in the general case. Assuming the convergence objective can be met, a desired augmentation to the objective is statistical efficiency on the way to convergence. A strategy for convergence in a streaming scenario would necessarily re-evaluate the model; HATT improvises a split evaluation heuristic more conducive to statistical efficiency and eventual convergence than to maximizing confidence in historical splits as does HoeffdingTree.
	
	As previously mentioned, in the classification setting, decision trees are grown by repetitively dividing the instance space and assigning a class (and possibly a probability distribution over classes) to each subspace. Where all training instances are available at the outset, as in batch learning, decision trees evaluate all available features (attributes) in order to decide which attribute should first be utilized to divide the space. According to the split heuristics we have discussed, the best attribute is one that maximizes increase in class purity based on some heuristic measuring class purity such as Information Gain or the Gini Coefficient. For example, if we have available the nominal features ``can-swim'' and ``is-a-mammal'', the latter feature would perfectly classify sharks and dolphins, splitting the instance space into two parts assuming no noise in the data, while the former attribute provides no information, as both sharks and dolphins  swim. Now suppose we wish to further classify the creatures as certain shark or dolphin species; we may use other attributes to recursively split the instance space.
	
	Where all training instances are not available at the outset, as in the case of online learning, it becomes necessary to determine when and how split the instance space. Hoeffding Tree addresses the problem by aiming to attain statistical confidence (using the Hoeffding Test) in the top attribute outperforming the second best attribute at each split decision. However, attaining this confidence is time-consuming. HATT utilizes information available before attaining this level of confidence by using the Hoeffding Test to determine whether a candidate attribute is better than the current split, rather than the second best attribute. If there is currently no split and a candidate split attribute tests to be significantly better, it is split upon; this decision may be revised as necessary. The net result is that in the stationary ($i.i.d$) setting that Hoeffding Tree assumes, HATT converges to the ideal tree that would have been obtained in a hypothetical batch setting that processes an infinity of training examples at once.

	\subsection{Convergence}
	
	Hoeffding Tree probabilistically bounds deviation from the ideal batch tree. It guarantees that the expected ``extensional'' or ``intensional'' disagreement from a batch tree are each independently bound by $\frac{\delta}{p}$, where $\delta$ is a tolerance level and $p$ is the leaf probability-- the probability that an example will fall into a leaf at a given level. ``Extensional disagreement'' is defined as the probability that a pair of decision trees will produce different predictions for a given example, and ``intensional disagreement'' is the probability that the path of an example will differ on the two trees \cite{domingos2000mining}. It is assumed that these guarantees hold in the limit when trained on an infinite dataset denoted $DT_*$. $p$ is assumed to be constant across all levels for simplicity.
	
	Note that the guarantees will weaken significantly as the number of leaves  increase, tending to probability bound $1$ as tree size increases. Increasing the complexity and size of data streams such that a larger tree is required increases the chance of a greater deviation. This limits the utility of guarantees for Hoeffding Tree to lower dimensional scenarios, though the trees themselves may be highly useful.
	
	HATT on the other hand, converges in probability to the batch decision tree under the following assumptions:
	\begin{itemize}
		\item No two attributes will have identical information gain. This is a simplifying assumption to ensure that we can always split given enough examples, because $\epsilon$ is monotonically decreasing.
		\item The data are independently and identically distributed ($i.i.d$)
	\end{itemize}
	
	Given these assumptions, we list three useful properties of HATT that follow as lemmas; the proofs are presented in the ``Extremely Fast Decision Tree'' work that introduces HATT \cite{manapragada2018extremely}.
	
	\begin{lemma}\label{lem:reaches_same_root_split_as_HT}
		
		For any input stream $S$, HATT learned from $S$ will have the same split attribute at the root as HT learned from $S$ at the time HT splits the root node.
		
	\end{lemma}	
	
	\begin{lemma} \label{lem:same_root_split_as_DT_almost_surely}
		If the input stream $S$ is $i.i.d$, the split attribute $X_R^{HATT}$ at the root node of HATT converges in probability to the split attribute $X_R^{DT_*}$ used at the root node of $DT_*$. That is, as the number of examples grows large, the probability that HATT will have at the root a split $X_R^{HATT}$ that matches the split $X_R^{DT_*}$ at the root node of $DT_*$ goes to 1.
	\end{lemma}	
	\begin{lemma}\label{thm:converges_to_dt}
		If the input stream $S$ is $i.i.d$, Hoeffding AnyTime Tree converges to the asymptotic batch tree in probability.
		
	\end{lemma}

	\subsection{Time and Space Complexity}
	\label{sec:complexity}
	\textbf{Space Complexity:}As detailed in \cite{manapragada2018extremely}, on nominal data with $d$ attributes, $v$ values per attribute, and $c$ classes, HATT requires $O(dvc)$ memory to store node statistics at each node, as does HT \cite{domingos2000mining}. The worst case space complexity is $O(v^{d-1}dvc)$, because there may be a maximum of $(1-v^d)/(1-v)$ nodes due to geometric tree growth. The space complexity for HT is given as $O(ldvc)$ in \cite{domingos2000mining}, where $l$ is the current number of leaves; for HATT, the space complexity is $O(ndvc)$, where n is the current total number of nodes, including internal nodes. Because $l$ is $O(n)$, space complexity is equivalent for HATT and HT.
	
	In the ensemble setting, space complexity is simply multiplied by the maximum ensemble size $m$ to obtain $O(mndvc)$. Ensembles which use change detectors add to the space complexity by maintaining a window (this is independent of whether the base learner is HT or HATT). The change detector used in such ensembles in our experiments is ADWIN, the default MOA option, which maintains a self-adjusting window of prediction errors. ADWIN is parameterized  with a parameter M that determines maximum window size as $W = M \times \sum_{i=0}^M 2^i$. Each ADWIN instance uses $O(Mlog(W/M))$ memory \cite{bifet2007learning}, and one instance is used per ensemble component, so the space complexity added by ensembles that use ADWIN is $O(mMlog(W/M))$.

	\textbf{Time Complexity:} Again, as detailed in \cite{manapragada2018extremely}, the worst-case cost of both split evaluation and updating node statistics while processing an example are $O(dvc)$ for HT and $O(hdvc)$ for HATT, where $h$ is the maximum height of the tree. One leaf (for HT) or one path (for HATT) have to be evaluated for splits and have their node statistics updated. 
	
	In the ensemble setting, time complexity is also simply multiplied by maximum ensemble size $m$, giving us $O(mdvc)$ for HT and $O(mhdvc)$ for HATT for both split evaluation and updating node statistics. Ensembles that use ADWIN add an additional amortized cost of $O(1)$ and worst case cost of $O(W)$ for processing each example per ADWIN instance, which evaluates to $O(m)$ amortized and $O(mW)$ worst case, as one ADWIN instance is used per ensemble component.
	
	\subsection{HATT in the context of concept drift}
	\label{sec:hattdrift}
	
	Hoeffding Tree is surprisingly responsive to concept drift given that it is designed for learning from stationary distributions and has no capacity to revise internal nodes once they have been added to the model. We posit that a major reason for this responsiveness is that when a split is created, the new leaves that are created start with no memory of the examples seen by the learner previously. As a result, if a split occurs after drift, the new leaves start with a fresh slate and learns the new distribution \cite{manapragada2020emergent}.
	
	HATT has the same property of new leaves starting with a fresh slate. However, it has the additional properties of---
	\begin{enumerate}
		\item forming new branches more readily (when it has evidence that a specific branch is better than  none, rather than better than any potential alternative) and
		\item replacing internal nodes with new splits when the new split becomes better than the existing one.
	\end{enumerate}
	The first of these properties enhances the speed with which HATT adjusts to drift, as it will develop new leaves reflecting a new distribution more rapidly. However, the second has both positive and negative aspects in the context of drift. Replacing an internal node can potentially remove a large section of the model that reflects the old distribution and allow it to be replaced by a new subtree that will grow to reflect the new distribution. However, the closer to the root a node is, the more evidence of older distributions will be retained in its node statistics. Hence, the closer to the root a node is, the longer it is likely to take for the evidence to grow sufficient to replace it. As a result, by the time an internal node is replaced following a drift, the subtrees below it are likely to have already had time to adjust to the new distribution, and hence replacing them may be detrimental in the near term until a new tree can be grown to replace them.
	
	\subsection{HATT in the Ensemble Setting}
	\label{sec:hattensemble}
	
	The choice of base predictor is important and Hoeffding Tree has been the mainstay of online ensemble learning. It has been argued that uncorrelated predictions are important for error reduction effect \cite{breiman1996bagging}; consequently that learner diversity is key to uncorrelated predictions \cite{kuncheva2003elusive}, implying that unstable learners are most suited as base learners for ensembles so small changes to the stream can cause significant changes to the base models, creating diversity in the ensemble. However, Hoeffding Tree is a stable learner \cite{DBLP:conf/icdm/GomesRB19}; that is, being provided slightly different versions of a stream does not greatly alter the decision tree model produced, on account of the Hoeffding Test that is used to decide each split with statistical confidence. Hoeffding AnyTime Tree can be argued to be perfectly stable in the long run---no matter the sequence in which the input is provided, it will converge to the ideal predictor as $t \rightarrow \infty$---however, it is far less stable than Hoeffding Tree in the short term on account of its incidentally adaptive nature as a result of constantly readjusting split decisions. This short term reduction in stability in HATT is a plausible reason to expect more accurate ensembles when it is used as a base predictor in ensembles in place of Hoeffding Tree.

	\section{Experimental Setup}
	\label{expsetup}

	We work with the MOA framework \cite{bifet2010moa}, which provides implementations of common ensembling strategies for decision trees, including Hoeffding Tree. Each ensemble method comprises either a bagging or boosting component, may involve a change detector and may also weight instances. Change detectors are usually used to determine whether to remove the worst performing trees in order to replace them with a new one if a significant increase in the error is observed.
	
	We use a testbench that to our best knowledge is the largest and most comprehensive in the literature, though there is ample room for improvement. We test on UCI datasets that are mostly drawn from real data as well as on synthetic streams with concept drift. We carefully selected twenty real datasets: two electricity datasets and the airlines and AWS dataset that are widely used in the concept drift literature, and as for the rest, the largest datasets from the UCI repository that involved an obvious classification task and had no missing values in order to reduce the confounding factor of how the algorithms handle missing values. We omitted the physics simulation datasets Hepmass, SUSY and Higgs as these are synthetic datasets with low information \cite{manapragada2018extremely}. The price variable in the AWS dataset was discretized into ten equal buckets and set as the classification target; similarly, the CO concentration variable in the sensor-CO dataset was discretized into 5 equal buckets and set as the classification target. The datasets we use and their key features are listed in Table \ref{table:datasets}.
	
	\begin{table}[h!]
		\centering
		\caption{Properties of Real Datasets}
		\begin{center}
\begin{tabular}{| c | l | m{1.5cm} | m{1.8cm} | c |}
	\hline
 & Dataset & Instances & Attributes (Numeric, Nominal) & Classes  \\ 
 	\hline
1 & airlines \cite{datasetairlines} & 539383 & 8 (3, 5) & 2 \\ 
2 & aws---price-discretized \cite{datasetaws-discrete} & 27410309 & 7 (4, 3) & 10 \\ 
3 & chess \cite{Dua:2019} & 28056 & 6(3, 3) & 18 \\ 
4 & covtype \cite{datasetcovtype, Dua:2019} & 581012 & 54 (10, 44) & 7 \\ 
5 & cpe \cite{datasetcpe, bifet2009new} & 1455525 & 72 (22, 50) & 10 \\ 
6 & fonts \cite{datasetfonts, Dua:2019} & 745000 & 411 (410, 1) & 153 \\ 
7 & hhar \cite{Stisen:2015:SDD:2809695.2809718, Dua:2019} & 43930257 & 9 (6, 3) & 6 \\ 
8 & kdd \cite{Dua:2019} & 4000000 & 42 (34, 8) & 23 \\ 
9 & localization \cite{datasetlocalization, Dua:2019} & 164860 & 8 (4, 4) & 11 \\ 
10 & miniboone \cite{datasetminiboone, Dua:2019} & 130065 & 50 (50, 0) & 2 \\ 
11 & nbaiot \cite{datasetnbaiot, Dua:2019} & 7062606 & 115 (115, 0) & 11 \\ 
12 & nswelec \cite{datasetelec, Dua:2019} & 45312 & 9 (7, 2) & 2 \\ 
13 & pamap2 \cite{datasetpamap, Dua:2019} & 3850505 & 53 (53,0) & 25 \\ 
14 & poker \cite{Dua:2019} & 1025010 & 10 (5, 5) & 10 \\ 
15 & pucrio \cite{datasetpucrio, Dua:2019} & 165632 & 18 (15, 3) & 5 \\ 
16 & sensor---home-activity \cite{datasetsensor, Dua:2019} & 919438 & 11 (11, 0) & 3 \\ 
17 & sensor---CO-discretized \cite{datasetgassensor2019discretized1, datasetgassensor2019discretized2, Dua:2019} & 4095000 & 19 (19, 0) & 5 \\ 
18 & skin \cite{datasetskin, Dua:2019} & 245057 & 3 (3, 0) & 2 \\ 
19 & tnelec \cite{Dua:2019} & 45781 & 4 (2, 2) & 20 \\ 
20 & wisdm \cite{datasetwisdm, Dua:2019} & 15630426 & 44 (43, 1) & 6 \\
	\hline
\end{tabular}
\end{center} 

		\label{table:datasets}
	\end{table}	
	
	UCI datasets are often ordered on some basis. As such, when processed sequentially in their native order, they provide a non-stationary data stream. We use them in this way. To assess performance on real world data in the absence of concept drift, we also use shuffled versions of each dataset. This ensures a stationary distribution. Our shuffled runs on real data use 10 randomised shuffles of each stream with fixed seeds so the experiments are reproducible.
	
	In order to assess ensembling strategies with VFDT and EFDT as base learners under high levels of concept drift, we chose a synthetic testbench that demonstrates noticeable differences in performance across different parametrizations. These synthetic streams and their parametrizations are listed in Table \ref{table:syntheticStreams}.

	\begin{table}[h!]
		\centering
		\scriptsize
		\caption{Synthetic Datasets}
		\makebox[1.0\textwidth]{
			\scalebox{0.9}{	
				
				\begin{tabular}{|c|p{11.8cm}|l|}
					\hline \\
					
					& MOA Stream & Shorthand  \\ 
					\hline
					
					1 & -s (RecurrentConceptDriftStream -x 200000 -y 200000 -z 100 -s (generators.AgrawalGenerator -f 2 -i 2) -d (generators.AgrawalGenerator -f 3 -i 3)) & recurrent---agrawal \\
2 & -s (RecurrentConceptDriftStream -x 200000 -y 200000 -z 100 -s (generators.LEDGenerator -i 2) -d (generators.LEDGeneratorDrift -i 3 -d 7)) & recurrent---led \\
3 & -s (RecurrentConceptDriftStream -x 200000 -y 200000 -z 100 -s (generators.RandomTreeGenerator -r 1 -i 1) -d (generators.RandomTreeGenerator -r 2 -i 2)) & recurrent---randomtree \\
4 & -s (RecurrentConceptDriftStream -x 200000 -y 200000 -z 100 -s (generators.SEAGenerator -f 2 -i 2) -d (generators.SEAGenerator -f 3 -i 3)) & recurrent---sea \\
5 & -s (RecurrentConceptDriftStream -x 200000 -y 200000 -z 100 -s (generators.STAGGERGenerator -i 2 -f 2) -d (generators.STAGGERGenerator -i 3 -f 3)) & recurrent---stagger \\
6 & -s (RecurrentConceptDriftStream -x 200000 -y 200000 -z 100 -s (generators.WaveformGenerator -i 2 -n) -d (generators.WaveformGeneratorDrift -i 3 -d 40 -n)) & recurrent---waveform \\
7 & -s (generators.HyperplaneGenerator -k 10 -t 0.0001 -i 2) & hyperplane---1 \\
8 & -s (generators.HyperplaneGenerator -k 10 -t 0.001 -i 2) & hyperplane---2 \\
9 & -s (generators.HyperplaneGenerator -k 5 -t 0.0001 -i 2) & hyperplane---3 \\
10 & -s (generators.HyperplaneGenerator -k 5 -t 0.001 -i 2) & hyperplane---4 \\
11 & -s (generators.RandomRBFGeneratorDrift -s 0.0001 -k 10 -i 2 -r 2) & rbf---drift-1 \\
12 & -s (generators.RandomRBFGeneratorDrift -s 0.0001 -k 50 -i 2 -r 2) & rbf---drift-2 \\
13 & -s (generators.RandomRBFGeneratorDrift -s 0.001 -k 10 -i 2 -r 2) & rbf---drift-3 \\
14 & -s (generators.RandomRBFGeneratorDrift -s 0.001 -k 50 -i 2 -r 2) & rbf---drift-4 \\
15 & -s (generators.monash.AbruptDriftGenerator -c  -o 1.0 -z 2 -n 2 -v 2 -r 2 -b 200000 -d Recurrent) & recurrent---abrupt---222 \\
16 & -s (generators.monash.AbruptDriftGenerator -c  -o 1.0 -z 3 -n 2 -v 2 -r 2 -b 200000 -d Recurrent) & recurrent---abrupt---322 \\
17 & -s (generators.monash.AbruptDriftGenerator -c  -o 1.0 -z 3 -n 3 -v 2 -r 2 -b 200000 -d Recurrent) & recurrent---abrupt---332 \\
18 & -s (generators.monash.AbruptDriftGenerator -c  -o 1.0 -z 3 -n 3 -v 3 -r 2 -b 200000 -d Recurrent) & recurrent---abrupt---333 \\
19 & -s (generators.monash.AbruptDriftGenerator -c  -o 1.0 -z 3 -n 3 -v 4 -r 2 -b 200000 -d Recurrent) & recurrent---abrupt---334 \\
20 & -s (generators.monash.AbruptDriftGenerator -c  -o 1.0 -z 3 -n 3 -v 5 -r 2 -b 200000 -d Recurrent) & recurrent---abrupt---335 \\
21 & -s (generators.monash.AbruptDriftGenerator -c  -o 1.0 -z 4 -n 2 -v 2 -r 2 -b 200000 -d Recurrent) & recurrent---abrupt---422 \\
22 & -s (generators.monash.AbruptDriftGenerator -c  -o 1.0 -z 4 -n 4 -v 4 -r 2 -b 200000 -d Recurrent) & recurrent---abrupt---444 \\
23 & -s (generators.monash.AbruptDriftGenerator -c  -o 1.0 -z 5 -n 2 -v 2 -r 2 -b 200000 -d Recurrent) & recurrent---abrupt---522 \\
24 & -s (generators.monash.AbruptDriftGenerator -c  -o 1.0 -z 5 -n 5 -v 5 -r 2 -b 200000 -d Recurrent) & recurrent---abrupt---555 \\

					\hline
				\end{tabular}
			}
		}
		\label{table:syntheticStreams}
	\end{table}

	Hyperplane \cite{hulten2001mining} and Radial Basis Function (RBF) \cite{bifet2009new} generators draw examples from a naturally evolving concept and allow exploration of the interactions between varying rates of drift and dimensionalities. 
	
	Our Recurrent AbruptDrift generator tweaks the AbruptDrift generator from \cite{webb2016characterizing} to add the option of generating recurrent abrupt drifts. It models a full conditional probability table for the target distribution $p_{Y|X}$ that grows exponentially in the number of input variables. Learning problems generated by this generator become significantly harder with small increases in dimensionality, possibly because of the relative independence of the variables. The suffix notation in the shorthand ``recurrent---abrupt---522'' used in tables conveys that that particular stream has 5 classes, 2 nominal attributes, and 2 values per attribute.  
	
	We use the RecurrentConceptDriftStream generator \cite{bifet2010moa} to generate drift between different parametrizations of synthetic streams commonly used in the concept drift literature that do not have in-built drift generation; these are the  Agrawal \cite{agrawal1992mining}, LED \cite{breiman1984classification}, RandomTree \cite{bifet2010moa}, SEA \cite{street2001streaming}, STAGGER \cite{Schlimmer1986}, and Waveform \cite{breiman1984classification} generators.
	
	All synthetic streams generate one million examples. Each synthetic stream generator is run with 10 differently initialized random seeds. Prequential accuracy is a curve that plots error at every timestep, so the results are reported thus: a measure of the mean error is obtained by averaging error across seeded runs for each epoch, and then averaging over the epoch-wise error averages. A measure of variance is obtained by computing the variance of the error across the seeded runs for each epoch, then averaging the epoch-wise variances. Note that the error reported for each epoch is itself the average across a thousand examples, which is the default for the evaluator in MOA.
	
	All ensembles were requisitioned with ten trees each. This ensemble size was close to the upper limit on what was feasible on a fairly large university cluster that made available 60 CPUs at a time for parallel workloads. Ensembles were otherwise parameterized with their MOA defaults. The boosting ensembles we use are: OzaBoost \cite{oza05}, OzaBoost with the ADWIN change detector (OzaBoostADWIN) \cite{bifet2007learning}, OnlineSmoothBoost \cite{servedio2003smooth, chen2012online}, Adaptable Diversity-based Online Boosting (ADOB) \cite{Santos2014}, and Boosting-Like Online Ensemble (BOLE) \cite{Barros2016}. The bagging ensembles in our experiments are: OzaBag \cite{oza05}, OzaBag with ADWIN (OzaBagADWIN), Leveraging Bagging (LevBag) \cite{bifet2010leveraging}, Leveraging Bagging without ADWIN (LevBagNoADWIN), and Adaptive Random Forest \cite{gomes2017adaptive}. Note that OzaBag and OzaBoost are two state-of-the-art online bagging and boosting algorithms---their core mechanisms extending bagging and boosting to online scenarios underpin all the other MOA ensembles that perform online bagging or boosting. We also compare just the individual EFDT and VFDT learners.
	
	All trees use NBAdaptive prediction at the leaves \cite{bifet2010moa}, that is, they either use Naive Bayes or Majority Class for prediction depending on whichever has been more accurate overall --- the cumulative accuracy of each approach from the beginning of the learning process at each leaf is recorded in order to facilitate this switching behavior.
	
	Note that variance is reported in Sections \ref{sec:shuffled} and \ref{sec:synthetic}, because these report error averages over 10 seeded runs for shuffled UCI streams and synthetic streams respectively, but not in Section \ref{sec:standard} where standard UCI streams are used unshuffled. Tables \ref{mevreal4}, \ref{mevrealshuf4}, and \ref{mevsyn4} do not present leaf counts as the MOA result files do not contain this information.
	
	Our implementations of EFDT and VFDT differ only in their split selection (and reselection) strategy; there are no implementation details that are otherwise different. All our experimental results and scripts are available for replicability at \url{github.com/chaitanya-m} in the \textit{exp\_analysis}, \textit{results2},  \textit{moa\_experiment\_scripts (ensembles branch)} repositories; code for VFDT, EFDT, and the Recurrent AbruptDrift generator is at \url{bitbucket.org/chaitm} in the \textit{driftgen} \textit{(thesis branch)} repository. 
	
	We use prequential accuracy as the primary evaluation measure. While the instantaneous feedback provided to the learner in prequential evaluation does not reflect a common setup in applied incremental learning, none of the learners in the study relies on or exploits it. All learners examined are general incremental learners, capable at any point in time of either updating the current model by learning from an example or applying that model to classify an example.  The prequential evaluation strategy is simply a convenient incremental setting for evaluating such systems. The critical feature of prequential learning is the it ensures that a learner always learns from an example before classifying it. We use prequential evaluation because it is the de facto standard for evaluation in the field of research.
	

	\FloatBarrier

	\section{Experimental Results}
	\label{expresults}

	In order to compare the relative prequential accuracy of EFDT as a base learner with the prequential accuracy of VFDT as a base learner, we tabulated their wins (datasets for which each achieves lower error) by ensemble and class of streams (UCI unshuffled, UCI shuffled 10-seed, Synthetic 10-seed). For each ensemble technique and class of stream we present the outcome of a one-tailed binomial test to determine the probability that the ensemble with EFDT as base learner would achieve so many wins if wins and losses were equiprobable (represented by the p-value). We use a statistical significance level of $\alpha=0.05$. The confidence interval represents the 95\% spread for probability of success for the EFDT-based ensemble in the comparison. To address potential concerns about multiple testing, we also for each set of experiments present the outcome of a Holm-Bonferroni correction.

	\subsection{UCI Streams}
	As explained above, following the experimental method of \cite{manapragada2018extremely}, we first use UCI datasets in their original order to assess natural datasets with inherent drift.  We then shuffle the same datasets, in order to assess performance when an input stream comes from a stationary distribution.
	
	\subsubsection{Standard UCI streams}
	\label{sec:standard}
	
	Tables \ref{mevreal0}--\ref{mevreal10} show the effect due to EFDT on 20 UCI data streams, which are largely drawn from real data. The prequential accuracy performance matches expectations; there is minimal complex drift in these scenarios, mostly arising from concatenating files together, as many streams were made available in files that were each associated with a single class. Under these conditions, class boundaries change periodically, an instance of a trivial concept drift. 
	
	Five of these tables show bagging ensembles, which favor EFDT over VFDT. Table \ref{mevreal0} shows the performance of vanilla OzaBag, in which EFDT outperforms VFDT on 16 streams, underperforms on 2 streams, and draws level on the remaining 2 streams. The p-value noted is 0.00066, well within a standard significance level of 0.05. 
	
\begin{table}[!ht]
	
	\caption{\label{mevreal0} OzaBag - UCI streams processed in original order
	} 
	\scriptsize
	\centering
		\begin{tabular}{p{6cm}?P{1.3cm}?P{1.3cm}}
			\toprule
				\multicolumn{1}{c?}{\textbf{Streams}} &  \multicolumn{1}{m{2.6cm}}{\textbf{VFDT Base}} & \multicolumn{1}{m{2.6cm}}{\textbf{EFDT Base}}\\
				\multicolumn{1}{c?}{} & Error & Error
				\\
				\midrule
				airlines & 0.34988 & \textbf{0.34133} \\
aws---price-discretized & 0.1469 & \textbf{0.14027} \\
chess & 0.57514 & \textbf{0.25848} \\
covertype & 0.14551 & \textbf{0.11684} \\
covpokelec & 0.146 & \textbf{0.13372} \\
fonts & 0.01729 & \textbf{0.00245} \\
hhar & 0.02375 & \textbf{0.00115} \\
kdd & 0.00088 & \textbf{0.00086} \\
localization & 0.09551 & \textbf{0.06576} \\
miniboone & \textit{\textbf{8e-05}} & \textit{\textbf{8e-05}} \\
nbaiot & 0.00522 & \textbf{0.00091} \\
nswelec & 0.17367 & \textbf{0.17226} \\
pamap2 & 0.04046 & \textbf{0.01335} \\
poker & 0.1613 & \textbf{0.15943} \\
pucrio & 0.02202 & \textbf{0.00089} \\
sensor---home-activity & 0.00208 & \textbf{0.00049} \\
sensor---CO-discretized & \textbf{0.01362} & 0.03894 \\
skin & \textit{\textbf{0.00062}} & \textit{\textbf{0.00062}} \\
tnelec & \textbf{0.0025} & 0.00528 \\
wisdm & 0.11476 & \textbf{0.0812} \\
\midrule
\textbf{Unique Wins} & \multicolumn{1}{l?}{\textbf{2}} & \multicolumn{1}{l}{\textbf{16}} \\
\midrule\multicolumn{3}{l}{A \textbf{bold} error value indicates higher accuracy, and \textit{\textbf{bold italics}} indicate a tie.}\\
\multicolumn{3}{l}{\textbf{One-tailed binomial test statistics}:  \textbf{p-value: 0.00066};   \textbf{Confidence Interval:  0.68974 --- 1} }\\

				\bottomrule
		\end{tabular}
\end{table}

\begin{table}[!ht]
	
	\caption{\label{mevreal1} OzaBagADWIN - UCI streams processed in original order
	} 
	\scriptsize
	\centering
	\makebox[0.6\textwidth]{
		\begin{tabular}{p{6cm}?P{1.3cm}?P{1.3cm}}
			\toprule
			\multicolumn{1}{c?}{\textbf{Streams}} &  \multicolumn{1}{m{2.6cm}}{\textbf{VFDT Base}} & \multicolumn{1}{m{2.6cm}}{\textbf{EFDT Base}}\\
			\multicolumn{1}{c?}{} & Error & Error
			\\
			\midrule
			airlines & \textbf{0.33446} & 0.33988 \\
aws---price-discretized & 0.14711 & \textbf{0.14057} \\
chess & 0.01741 & \textbf{0.01379} \\
covertype & 0.15069 & \textbf{0.10075} \\
covpokelec & 0.2132 & \textbf{0.18577} \\
fonts & 0.00128 & \textbf{8e-04} \\
hhar & 0.0041 & \textbf{0.00093} \\
kdd & \textbf{0.00047} & 0.00055 \\
localization & 0.06409 & \textbf{0.04699} \\
miniboone & \textit{\textbf{8e-05}} & \textit{\textbf{8e-05}} \\
nbaiot & 0.00028 & \textbf{0.00014} \\
nswelec & 0.16307 & \textbf{0.12798} \\
pamap2 & 0.00858 & \textbf{0.00282} \\
poker & 0.25234 & \textbf{0.24115} \\
pucrio & 0.00092 & \textbf{0.00066} \\
sensor---home-activity & 3e-04 & \textbf{0.00029} \\
sensor---CO-discretized & 0.02445 & \textbf{0.02192} \\
skin & \textit{\textbf{0.00036}} & \textit{\textbf{0.00036}} \\
tnelec & \textbf{0.00278} & 0.00343 \\
wisdm & 0.10881 & \textbf{0.07696} \\
\midrule
\textbf{Unique Wins} & \multicolumn{1}{l?}{\textbf{3}} & \multicolumn{1}{l}{\textbf{15}} \\
\midrule\multicolumn{3}{l}{A \textbf{bold} error value indicates higher accuracy, and \textit{\textbf{bold italics}} indicate a tie.}\\
\multicolumn{3}{l}{\textbf{One-tailed binomial test statistics}:  \textbf{p-value: 0.00377};   \textbf{Confidence Interval:  0.62332 --- 1} }\\

			\bottomrule
		\end{tabular}}
\end{table}

\begin{table}[!ht]
	
	\caption{\label{mevreal2} LevBagNoADWIN - UCI streams processed in original order
	} 
	\scriptsize
	\centering
	\makebox[0.6\textwidth]{
		\begin{tabular}{p{6cm}?P{1.3cm}?P{1.3cm}}
			\toprule
			\multicolumn{1}{c?}{\textbf{Streams}} &  \multicolumn{1}{m{2.6cm}}{\textbf{VFDT Base}} & \multicolumn{1}{m{2.6cm}}{\textbf{EFDT Base}}\\
			\multicolumn{1}{c?}{} & Error & Error
			\\
			\midrule
				airlines & 0.35124 & \textbf{0.34829} \\
aws---price-discretized & 0.13042 & \textbf{0.12233} \\
chess & 0.48134 & \textbf{0.18238} \\
covertype & 0.099 & \textbf{0.07295} \\
covpokelec & \textbf{0.06218} & 0.08186 \\
fonts & 0.01641 & \textbf{0.00156} \\
hhar & 0.01012 & \textbf{0.00109} \\
kdd & \textbf{0.00073} & 0.00075 \\
localization & 0.05593 & \textbf{0.05218} \\
miniboone & \textbf{4e-05} & 5e-05 \\
nbaiot & 0.00089 & \textbf{0.00038} \\
nswelec & 0.15643 & \textbf{0.14809} \\
pamap2 & 0.15539 & \textbf{0.02625} \\
poker & \textbf{0.05333} & 0.09355 \\
pucrio & 0.01239 & \textbf{0.00115} \\
sensor---home-activity & 0.0029 & \textbf{0.00071} \\
sensor---CO-discretized & \textbf{0.01437} & 0.03077 \\
skin & \textbf{0.00015} & 0.00016 \\
tnelec & \textit{\textbf{0.00324}} & \textit{\textbf{0.00324}} \\
wisdm & 0.08572 & \textbf{0.06601} \\
\midrule
\textbf{Unique Wins} & \multicolumn{1}{l?}{\textbf{6}} & \multicolumn{1}{l}{\textbf{13}} \\
\midrule\multicolumn{3}{l}{A \textbf{bold} error value indicates higher accuracy, and \textit{\textbf{bold italics}} indicate a tie.}\\
\multicolumn{3}{l}{\textbf{One-tailed binomial test statistics}:  \textbf{p-value: 0.08353};   \textbf{Confidence Interval:  0.47003 --- 1} }\\

				\bottomrule
		\end{tabular}}
\end{table}

\begin{table}[!ht]
	
	\caption{\label{mevreal3} LevBag - UCI streams processed in original order
	} 
	\scriptsize
	\centering
	\makebox[0.6\textwidth]{
		\begin{tabular}{p{6cm}?P{1.3cm}?P{1.3cm}}
			\toprule
			\multicolumn{1}{c?}{\textbf{Streams}} &  \multicolumn{1}{m{2.6cm}}{\textbf{VFDT Base}} & \multicolumn{1}{m{2.6cm}}{\textbf{EFDT Base}}\\
			\multicolumn{1}{c?}{} & Error & Error
			\\
			\midrule
				airlines & \textbf{0.34316} & 0.34607 \\
aws---price-discretized & 0.13038 & \textbf{0.12272} \\
chess & 0.01576 & \textbf{0.01355} \\
covertype & 0.09768 & \textbf{0.06784} \\
covpokelec & 0.17382 & \textbf{0.13015} \\
fonts & 0.00114 & \textbf{0.00067} \\
hhar & 0.00088 & \textbf{0.00063} \\
kdd & \textbf{0.00034} & 0.00037 \\
localization & 0.03834 & \textbf{0.03781} \\
miniboone & \textbf{4e-05} & 5e-05 \\
nbaiot & 9e-05 & \textbf{8e-05} \\
nswelec & 0.11276 & \textbf{0.10457} \\
pamap2 & 0.003 & \textbf{0.00138} \\
poker & 0.222 & \textbf{0.1684} \\
pucrio & \textbf{0.00101} & 0.00103 \\
sensor---home-activity & 3e-04 & \textbf{0.00025} \\
sensor---CO-discretized & \textbf{0.01412} & 0.0142 \\
skin & \textit{\textbf{0.00014}} & \textit{\textbf{0.00014}} \\
tnelec & \textit{\textbf{0.00267}} & \textit{\textbf{0.00267}} \\
wisdm & 0.10197 & \textbf{0.06728} \\
\midrule
\textbf{Unique Wins} & \multicolumn{1}{l?}{\textbf{5}} & \multicolumn{1}{l}{\textbf{13}} \\
\midrule\multicolumn{3}{l}{A \textbf{bold} error value indicates higher accuracy, and \textit{\textbf{bold italics}} indicate a tie.}\\
\multicolumn{3}{l}{\textbf{One-tailed binomial test statistics}:  \textbf{p-value: 0.04813};   \textbf{Confidence Interval:  0.50217 --- 1} }\\

				\bottomrule
		\end{tabular}}
	
\end{table}

\begin{table}[!ht]
	
	\caption{\label{mevreal4} Adaptive Random Forest - UCI streams processed in original order
	} 
	\scriptsize
	\centering
	\makebox[0.6\textwidth]{
		\begin{tabular}{p{6cm}?P{1.3cm}?P{1.3cm}}
			\toprule
			\multicolumn{1}{c?}{\textbf{Streams}} &  \multicolumn{1}{m{2.6cm}}{\textbf{VFDT Base}} & \multicolumn{1}{m{2.6cm}}{\textbf{EFDT Base}}\\
			\multicolumn{1}{c?}{} & Error & Error 
			\\
			\midrule
				airlines & 0.34994 & \textbf{0.33776} \\
aws---price-discretized & \textbf{0.18187} & 0.19474 \\
chess & 0.01924 & \textbf{0.0139} \\
covertype & 0.11075 & \textbf{0.07057} \\
covpokelec & 0.18964 & \textbf{0.15533} \\
fonts & 0.006 & \textbf{0.00599} \\
hhar & 0.0015 & \textbf{0.00075} \\
kdd & 0.00044 & \textbf{0.00043} \\
localization & \textbf{0.03333} & 0.03576 \\
miniboone & \textbf{4e-05} & 5e-05 \\
nbaiot & \textit{\textbf{4e-05}} & \textit{\textbf{4e-05}} \\
nswelec & 0.13567 & \textbf{0.11857} \\
pamap2 & 0.00151 & \textbf{0.00113} \\
poker & 0.24551 & \textbf{0.20895} \\
pucrio & \textit{\textbf{0.00044}} & \textit{\textbf{0.00044}} \\
sensor---home-activity & 5e-04 & \textbf{0.00024} \\
sensor---CO-discretized & 0.00834 & \textbf{0.00765} \\
skin & 0.00013 & \textbf{0.00011} \\
tnelec & 0.01257 & \textbf{0.00748} \\
wisdm & 0.10885 & \textbf{0.07728} \\
\midrule
\textbf{Unique Wins} & \multicolumn{1}{l?}{\textbf{3}} & \multicolumn{1}{l}{\textbf{15}} \\
\midrule\multicolumn{3}{l}{A \textbf{bold} error value indicates higher accuracy, and \textit{\textbf{bold italics}} indicate a tie.}\\
\multicolumn{3}{l}{\textbf{One-tailed binomial test statistics}:  \textbf{p-value: 0.00377};   \textbf{Confidence Interval:  0.62332 --- 1} }\\

				\bottomrule
		\end{tabular}}
\end{table}

	
\begin{table}[!ht]
	
	\caption{\label{mevreal5} OzaBoost - UCI streams processed in original order
	} 
	\scriptsize
	\centering
	\makebox[0.6\textwidth]{
		\begin{tabular}{p{6cm}?P{1.3cm}?P{1.3cm}}
			\toprule
			\multicolumn{1}{c?}{\textbf{Streams}} &  \multicolumn{1}{m{2.6cm}}{\textbf{VFDT Base}} & \multicolumn{1}{m{2.6cm}}{\textbf{EFDT Base}}\\
			\multicolumn{1}{c?}{} & Error & Error
			\\
			\midrule
				airlines & \textbf{0.36114} & 0.36415 \\
aws---price-discretized & 0.14541 & \textbf{0.13773} \\
chess & 0.84848 & \textbf{0.24224} \\
covertype & 0.08286 & \textbf{0.05701} \\
covpokelec & 0.09133 & \textbf{0.09009} \\
fonts & 0.00167 & \textbf{0.00166} \\
hhar & 0.00142 & \textbf{0.00103} \\
kdd & \textbf{0.00176} & 0.00183 \\
localization & 0.04061 & \textbf{0.03807} \\
miniboone & \textit{\textbf{9e-05}} & \textit{\textbf{9e-05}} \\
nbaiot & 0.0092 & \textbf{0.00124} \\
nswelec & 0.14113 & \textbf{0.13004} \\
pamap2 & 0.28883 & \textbf{0.00364} \\
poker & \textbf{0.10215} & 0.10375 \\
pucrio & \textbf{0.00532} & 0.02177 \\
sensor---home-activity & 0.00145 & \textbf{8e-04} \\
sensor---CO-discretized & \textbf{0.00385} & 0.01015 \\
skin & \textit{\textbf{0.20681}} & \textit{\textbf{0.20681}} \\
tnelec & \textbf{0.00148} & 0.00152 \\
wisdm & 0.10207 & \textbf{0.06448} \\
\midrule
\textbf{Unique Wins} & \multicolumn{1}{l?}{\textbf{6}} & \multicolumn{1}{l}{\textbf{12}} \\
\midrule\multicolumn{3}{l}{A \textbf{bold} error value indicates higher accuracy, and \textit{\textbf{bold italics}} indicate a tie.}\\
\multicolumn{3}{l}{\textbf{One-tailed binomial test statistics}:  \textbf{p-value: 0.11894};   \textbf{Confidence Interval:  0.44595 --- 1} }\\

				\bottomrule
		\end{tabular}}
\end{table}

\begin{table}[!ht]
	
	\caption{\label{mevreal6} OzaBoostADWIN - UCI streams processed in original order
	} 
	\scriptsize
	\centering
	\makebox[0.6\textwidth]{
		\begin{tabular}{p{6cm}?P{1.3cm}?P{1.3cm}}
			\toprule
			\multicolumn{1}{c?}{\textbf{Streams}} &  \multicolumn{1}{m{2.6cm}}{\textbf{VFDT Base}} & \multicolumn{1}{m{2.6cm}}{\textbf{EFDT Base}}\\
			\multicolumn{1}{c?}{} & Error & Error
			\\
			\midrule
				airlines & \textbf{0.38107} & 0.38181 \\
aws---price-discretized & 0.14548 & \textbf{0.13786} \\
chess & 0.18648 & \textbf{0.10631} \\
covertype & 0.08812 & \textbf{0.0856} \\
covpokelec & \textbf{0.12989} & 0.14332 \\
fonts & 0.42996 & \textbf{0.00853} \\
hhar & 0.14148 & \textbf{0.12517} \\
kdd & \textbf{0.2722} & 0.28153 \\
localization & 0.0501 & \textbf{0.04536} \\
miniboone & \textbf{6e-05} & 7e-05 \\
nbaiot & 0.43386 & \textbf{0.37867} \\
nswelec & 0.09989 & \textbf{0.0983} \\
pamap2 & 0.32267 & \textbf{0.24907} \\
poker & \textbf{0.14994} & 0.18364 \\
pucrio & \textbf{0.00124} & 0.2646 \\
sensor---home-activity & 0.32472 & \textbf{0.00088} \\
sensor---CO-discretized & \textbf{0.30825} & 0.34715 \\
skin & \textit{\textbf{0.99995}} & \textit{\textbf{0.99995}} \\
tnelec & \textbf{0.00143} & 0.00165 \\
wisdm & 0.08342 & \textbf{0.07332} \\
\midrule
\textbf{Unique Wins} & \multicolumn{1}{l?}{\textbf{8}} & \multicolumn{1}{l}{\textbf{11}} \\
\midrule\multicolumn{3}{l}{A \textbf{bold} error value indicates higher accuracy, and \textit{\textbf{bold italics}} indicate a tie.}\\
\multicolumn{3}{l}{\textbf{One-tailed binomial test statistics}:  \textbf{p-value: 0.3238};   \textbf{Confidence Interval:  0.36812 --- 1} }\\

				\bottomrule
		\end{tabular}}
\end{table}

	
\begin{table}[!ht]
	
	\caption{\label{mevreal7} ADOB - UCI streams processed in original order
	} 
	\scriptsize
	\centering
	\makebox[0.6\textwidth]{
		\begin{tabular}{p{6cm}?P{1.3cm}?P{1.3cm}}
			\toprule
			\multicolumn{1}{c?}{\textbf{Streams}} &  \multicolumn{1}{m{2.6cm}}{\textbf{VFDT Base}} & \multicolumn{1}{m{2.6cm}}{\textbf{EFDT Base}}\\
			\multicolumn{1}{c?}{} & Error & Error
			\\
			\midrule
				airlines & \textbf{0.36174} & 0.36964 \\
aws---price-discretized & 0.14263 & \textbf{0.13371} \\
chess & 0.60293 & \textbf{0.16659} \\
covertype & 0.07985 & \textbf{0.05784} \\
covpokelec & 0.09125 & \textbf{0.08324} \\
fonts & \textit{\textbf{0.00038}} & \textit{\textbf{0.00038}} \\
hhar & 0.00152 & \textbf{8e-04} \\
kdd & \textbf{0.00023} & 0.00026 \\
localization & 0.04004 & \textbf{0.03828} \\
miniboone & 0.00017 & \textbf{0.00016} \\
nbaiot & \textbf{0.00354} & 0.00688 \\
nswelec & 0.13748 & \textbf{0.13211} \\
pamap2 & 0.08868 & \textbf{0.00342} \\
poker & \textbf{0.0915} & 0.10831 \\
pucrio & 0.00289 & \textbf{0.00117} \\
sensor---home-activity & 0.00206 & \textbf{0.00056} \\
sensor---CO-discretized & \textbf{0.00295} & 0.00713 \\
skin & \textit{\textbf{0.00019}} & \textit{\textbf{0.00019}} \\
tnelec & 0.00172 & \textbf{0.00159} \\
wisdm & 0.09067 & \textbf{0.06247} \\
\midrule
\textbf{Unique Wins} & \multicolumn{1}{l?}{\textbf{5}} & \multicolumn{1}{l}{\textbf{13}} \\
\midrule\multicolumn{3}{l}{A \textbf{bold} error value indicates higher accuracy, and \textit{\textbf{bold italics}} indicate a tie.}\\
\multicolumn{3}{l}{\textbf{One-tailed binomial test statistics}:  \textbf{p-value: 0.04813};   \textbf{Confidence Interval:  0.50217 --- 1} }\\

				\bottomrule
		\end{tabular}}
\end{table}

\begin{table}[!ht]
	
	\caption{\label{mevreal8} BOLE - UCI streams processed in original order
	} 
	\scriptsize
	\centering
	\makebox[0.6\textwidth]{
		\begin{tabular}{p{6cm}?P{1.3cm}?P{1.3cm}}
			\toprule
			\multicolumn{1}{c?}{\textbf{Streams}} &  \multicolumn{1}{m{2.6cm}}{\textbf{VFDT Base}} & \multicolumn{1}{m{2.6cm}}{\textbf{EFDT Base}}\\
			\multicolumn{1}{c?}{} & Error & Error
			\\
			\midrule
				airlines & \textbf{0.36173} & 0.36963 \\
aws---price-discretized & 0.14263 & \textbf{0.13371} \\
chess & 0.45517 & \textbf{0.16648} \\
covertype & 0.07985 & \textbf{0.05773} \\
covpokelec & 0.09122 & \textbf{0.08315} \\
fonts & \textit{\textbf{0.00039}} & \textit{\textbf{0.00039}} \\
hhar & 0.00152 & \textbf{8e-04} \\
kdd & \textbf{0.00023} & 0.00026 \\
localization & 0.04005 & \textbf{0.0383} \\
miniboone & 0.00017 & \textbf{0.00016} \\
nbaiot & \textbf{0.00355} & 0.00688 \\
nswelec & 0.13737 & \textbf{0.132} \\
pamap2 & 0.04702 & \textbf{0.00342} \\
poker & \textbf{0.09138} & 0.10831 \\
pucrio & 0.00288 & \textbf{0.00116} \\
sensor---home-activity & 0.00207 & \textbf{0.00057} \\
sensor---CO-discretized & \textbf{0.00294} & 0.00713 \\
skin & \textit{\textbf{0.00017}} & \textit{\textbf{0.00017}} \\
tnelec & 0.00172 & \textbf{0.00159} \\
wisdm & 0.09066 & \textbf{0.06247} \\
\midrule
\textbf{Unique Wins} & \multicolumn{1}{l?}{\textbf{5}} & \multicolumn{1}{l}{\textbf{13}} \\
\midrule\multicolumn{3}{l}{A \textbf{bold} error value indicates higher accuracy, and \textit{\textbf{bold italics}} indicate a tie.}\\
\multicolumn{3}{l}{\textbf{One-tailed binomial test statistics}:  \textbf{p-value: 0.04813};   \textbf{Confidence Interval:  0.50217 --- 1} }\\

				\bottomrule
		\end{tabular}}
\end{table}

\begin{table}[!ht]
	
	\caption{\label{mevreal9} OnlineSmoothBoost - UCI streams processed in original order
	} 
	\scriptsize
	\centering
	\makebox[0.6\textwidth]{
		\begin{tabular}{p{6cm}?P{1.3cm}?P{1.3cm}}
			\toprule
			\multicolumn{1}{c?}{\textbf{Streams}} &  \multicolumn{1}{m{2.6cm}}{\textbf{VFDT Base}} & \multicolumn{1}{m{2.6cm}}{\textbf{EFDT Base}}\\
			\multicolumn{1}{c?}{} & Error & Error
			\\
			\midrule
			airlines & 0.34321 & \textbf{0.33784} \\
aws---price-discretized & 0.14576 & \textbf{0.13951} \\
chess & 0.47655 & \textbf{0.28121} \\
covertype & 0.15224 & \textbf{0.11009} \\
covpokelec & 0.13829 & \textbf{0.12604} \\
fonts & 0.01291 & \textbf{0.00228} \\
hhar & 0.01685 & \textbf{0.00245} \\
kdd & 0.00086 & \textbf{0.00076} \\
localization & 0.09631 & \textbf{0.06973} \\
miniboone & \textbf{0.00011} & 0.00013 \\
nbaiot & 0.00358 & \textbf{0.00097} \\
nswelec & 0.17422 & \textbf{0.16039} \\
pamap2 & 0.05008 & \textbf{0.03045} \\
poker & \textbf{0.13754} & 0.16386 \\
pucrio & 0.02314 & \textbf{0.00119} \\
sensor---home-activity & 0.00382 & \textbf{9e-04} \\
sensor---CO-discretized & \textbf{0.01256} & 0.03383 \\
skin & \textbf{0.00051} & 0.00055 \\
tnelec & \textbf{0.00237} & 0.00498 \\
wisdm & 0.12901 & \textbf{0.09535} \\
\midrule
\textbf{Unique Wins} & \multicolumn{1}{l?}{\textbf{5}} & \multicolumn{1}{l}{\textbf{15}} \\
\midrule\multicolumn{3}{l}{A \textbf{bold} error value indicates higher accuracy, and \textit{\textbf{bold italics}} indicate a tie.}\\
\multicolumn{3}{l}{\textbf{One-tailed binomial test statistics}:  \textbf{p-value: 0.02069};   \textbf{Confidence Interval:  0.54442 --- 1} }\\

			\bottomrule
		\end{tabular}}
\end{table}

\begin{table}[!ht]
	
	\caption{\label{mevreal10} Plain single learners (no ensemble) - UCI streams processed in original order
	} 
	\scriptsize
	\centering
	\makebox[0.6\textwidth]{
		\begin{tabular}{p{6cm}?P{1.3cm}?P{1.3cm}}
			\toprule
			\multicolumn{1}{c?}{\textbf{Streams}} &  \multicolumn{1}{m{2.6cm}}{\textbf{VFDT Base}} & \multicolumn{1}{m{2.6cm}}{\textbf{EFDT Base}}\\
			\multicolumn{1}{c?}{} & Error & Error
			\\
			\midrule
			airlines & 0.34955 & \textbf{0.34782} \\
aws---price-discretized & 0.14728 & \textbf{0.14143} \\
chess & 0.65807 & \textbf{0.31152} \\
covertype & 0.18905 & \textbf{0.15431} \\
covpokelec & \textbf{0.18665} & 0.19391 \\
fonts & 0.01781 & \textbf{0.003} \\
hhar & 0.03749 & \textbf{0.00357} \\
kdd & \textbf{0.00093} & 0.00094 \\
localization & 0.13299 & \textbf{0.09754} \\
miniboone & \textbf{0.00011} & 0.00013 \\
nbaiot & 0.00368 & \textbf{0.00076} \\
nswelec & 0.19861 & \textbf{0.19283} \\
pamap2 & 0.0972 & \textbf{0.06136} \\
poker & \textbf{0.20509} & 0.218 \\
pucrio & 0.01457 & \textbf{0.00147} \\
sensor---home-activity & 0.00906 & \textbf{0.00135} \\
sensor---CO-discretized & \textbf{0.02633} & 0.06655 \\
skin & \textbf{0.00052} & 0.00055 \\
tnelec & \textbf{0.00235} & 0.00496 \\
wisdm & 0.18796 & \textbf{0.14576} \\
\midrule
\textbf{Unique Wins} & \multicolumn{1}{l?}{\textbf{7}} & \multicolumn{1}{l}{\textbf{13}} \\
\midrule\multicolumn{3}{l}{A \textbf{bold} error value indicates higher accuracy, and \textit{\textbf{bold italics}} indicate a tie.}\\
\multicolumn{3}{l}{\textbf{One-tailed binomial test statistics}:  \textbf{p-value: 0.13159};   \textbf{Confidence Interval:  0.44197 --- 1} }\\

			\bottomrule
		\end{tabular}}
\end{table}			
	
	Table \ref{mevreal1} shows the performance of OzaBagADWIN, a version of OzaBag with the ADWIN change detector, which maintains a variable-length, self-adjusting window of observed error; when a change is detected in one of the base classifiers, it is replaced with a new one. This makes little difference to the performance of EFDT over VFDT, with outperformance on 15 streams (p-value 0.00377; this is also significant at a Holm-Bonferroni level of $0.05/9 = 0.00556$), potentially implying that in scenarios without significant levels of concept drift, ADWIN does not have an effect on EFDT bagging ensembles. However, as we discuss in Section \ref{sec:synthetic}, ADWIN interacts adversely with EFDT-based bagging ensembles when significant concept drift is present.

	Table \ref{mevreal2} shows the prequential accuracy performance comparison with Leveraging Bagging without ADWIN (LevBagNoADWIN), and Table \ref{mevreal3} shows the comparison for LeveragingBagging (LevBag). Leveraging Bagging \cite{bifet2010leveraging} is a variant of OzaBag that makes the critical observation that the original OzaBag algorithm parameterizes the Poisson distribution used for determining the weight of a sample with value 1---which causes a third of examples to be ignored as the value drawn from $Pois(\lambda=1)$ is 0 about $34\%$ of the time. Leveraging Bagging uses instead Poisson(6) which allows using many more examples, giving it a significant advantage in the streaming setting while retaining the bagging heuristic through differential weighting of examples. Leveraging Bagging (Table \ref{mevreal3}) performs better with EFDT than with VFDT---we obtain p-value of 0.04813, which is significant at the 0.05 significance level. The removal of the change detector causes a slight deterioration, with a p-value of 0.08353  that falls outside a significance level of 0.05 (Table \ref{mevreal2}), due to there being one more stream on which the EFDT Based system underperforms compared to the VFDT based system, taking the EFDT wins to 13 with 6 losses. Based on this observation and the observation concerning OzaBag, we do not find the role of ADWIN to be substantial in streams with minimal concept drift when bagging strategies are applied.
	
	Table \ref{mevreal4} shows the performance of Adaptive Random Forest \cite{gomes2017adaptive}, an adaptive, online variant of classic random forests. The EFDT based system registers a clear outperformance on prequential accuracy (p-value 0.00377).
	
	As for boosting ensembles, while EFDT still wins more often than it loses, neither OzaBoost (Table \ref{mevreal5}, 12 wins/6 losses, p-value 0.11894) nor its variant with ADWIN change detection (Table \ref{mevreal6}, 11 wins/8 losses, p-value 0.3238) benefit EFDT with significance. The general success of online bagging strategies over straightforward online boosting strategies is noted in \cite{bifet2010leveraging}, and while tangential to ensemble comparison, this discrepancy appears to work in favor of further setting apart EFDT based bagging ensembles in their outperformance.

	However, boosting regimes with greater particularity in weighting misclassified examples favor EFDT with significance. We observe this with ADOB (Table \ref{mevreal7}, p-value 0.04813), a variant of online boosting that rearranges ensemble components in increasing order of accuracy---misclassified examples are given half the total weight when passed to the next ensemble component in online boosting, and thus ADOB's strategy optimizes for learning misclassified instances sooner. BOLE (Table \ref{mevreal8}), a variant of ADOB which allows poorly performing ensemble members to vote, and OnlineSmoothBoost(Table \ref{mevreal9}), which provides a rationalized continuous weighting scheme for examples in contrast with the stepped Poisson weighting provided by OzaBoost, both also benefit EFDT with high significance, with p-values of 0.04813 and 0.02069 respectively. Note that we use ADOB and BOLE for their core strategies without change detectors (their MOA implementations default to wrapping ensemble components with change detectors, while the respective papers require wrapping the entire ensemble---this ambiguity was not of interest to our experimentation, but the core strategies were).
	
	Plain EFDT outperforms plain VFDT on 13 streams (Table \ref{mevreal10}), with VFDT registering lower prequential accuracy on the remaining 7.
	
	
	The Holm-Bonferroni adjusted p-values for a familywise significance level of 0.05 with 10 hypothesis tests (we have ten ensembles) are 0.005, 0.00556, 0.00625, 0.0071, 0.00833, 0.01, 0.0125, 0.01666, 0.025, and 0.05.
	
	Our ranked p-values are 0.00066, 0.00377, 0.00377, 0.02059, 0.04813, 0.04813, 0.04813, 0.08353, 0.11894, 0.3238. Only the first three alternate hypotheses would be rejected using a Holm-Bonferroni multiple testing correction, resulting in findings of significant improvements in the number of data streams for which the ensemble technique is more accurate using EFDT than VFDT for OzaBag, OzaBagAdwin and Adaptive Random Forest.
	
	In summary, as expected, the greater variance induced by EFDT's less conservative splitting mechanism proves productive when drift is not extreme. All types of ensemble learner achieve lower accuracy for more datasets than not when EFDT is used as a base learner in place of VFDT. The frequency of these wins is statistically significant at the 0.05 level for all ensemble techniques other than LeveragingBagging in the case that ADWIN is disabled (when the use of EFDT still achieves lower error for 13 datasets, but VFDT achieves lower error for 6, preventing a statistically significant win). When a Holm-Bonferroni correction is applied, three of the ten wins remain statistically significant.

	
	\subsubsection{Shuffled UCI streams}
	\label{sec:shuffled}

	We removed all concept drift from the data streams used in Section~\ref{sec:standard} so as to understand whether the advantage observed for streams containing drift is retained in the case of a static generating distribution. This was achieved through the simple expedient of randomly shuffling each stream before submitting it for prequential evaluation. The results on shuffled data streams are based on averaged prequential performance over 10 shuffles of each data stream, with fixed random seeds so the experiments can be replicated easily. These results are shown in Tables \ref{mevrealshuf0} through \ref{mevrealshuf10}.
	
	
\begin{table}
	
	\caption{\label{mevrealshuf0} OzaBag - Shuffled UCI streams
	} 
	\scriptsize
	\centering
	\makebox[0.6\textwidth]{
		\begin{tabular}{p{6cm}?P{1.3cm}?P{1.3cm}?P{1.3cm}?P{1.3cm}}
				\toprule
				\multicolumn{1}{c?}{\textbf{Streams}} &  \multicolumn{2}{m{2.9cm}}{\textbf{VFDT Base}} &
				\multicolumn{2}{m{2.9cm}}{\textbf{EFDT Base}}\\
				\multicolumn{1}{c?}{} & Error & Variance & Error & Variance
				\\
				\midrule
				airlines & \textbf{0.35662} & 0.00023 & 0.36085 & 0.00023 \\
aws---price-discretized & 0.14668 & 0.00013 & \textbf{0.14018} & 0.00012 \\
chess & 0.67124 & 0.00023 & \textbf{0.59518} & 3e-04 \\
covertype & 0.27185 & 0.00021 & \textbf{0.25813} & 2e-04 \\
covpokelec & \textbf{0.29823} & 0.00029 & 0.34712 & 0.00043 \\
fonts & \textit{\textbf{0.00075}} & 0 & \textit{\textbf{0.00075}} & 0 \\
hhar & 0.05203 & 6e-05 & \textbf{0.03088} & 3e-05 \\
kdd & 0.00098 & 0 & \textbf{8e-04} & 0 \\
localization & 0.35064 & 0.00023 & \textbf{0.32112} & 0.00068 \\
miniboone & 0.10228 & 0.00012 & \textbf{0.10025} & 0.00011 \\
nbaiot & 0.04758 & 0.0026 & \textbf{0.00216} & 3e-05 \\
nswelec & 0.23156 & 2e-04 & \textbf{0.23054} & 2e-04 \\
pamap2 & 0.1234 & 0.00016 & \textbf{0.10654} & 0.00014 \\
poker & \textbf{0.26286} & 0.00024 & 0.27569 & 0.00043 \\
pucrio & 0.1063 & 0.00018 & \textbf{0.04309} & 8e-05 \\
sensor---home-activity & 0.06297 & 9e-05 & \textbf{0.03715} & 0.00011 \\
sensor---CO-discretized & 0.18668 & 0.00022 & \textbf{0.17228} & 0.00022 \\
skin & 0.01661 & 2e-05 & \textbf{0.00973} & 1e-05 \\
tnelec & \textit{\textbf{0.00663}} & 1e-05 & \textit{\textbf{0.00663}} & 1e-05 \\
wisdm & 0.17683 & 0.00018 & \textbf{0.11814} & 0.00011 \\
\midrule
\textbf{Unique Wins} & \multicolumn{2}{l?}{\textbf{3}} & \multicolumn{2}{l}{\textbf{15}} \\
\midrule\multicolumn{5}{l}{A \textbf{bold} error value indicates higher accuracy, and \textit{\textbf{bold italics}} indicate a tie.}\\
\multicolumn{5}{l}{\textbf{One-tailed binomial test statistics}:  \textbf{p-value: 0.00377};   \textbf{Confidence Interval:  0.62332 --- 1} }\\

				\bottomrule
		\end{tabular}}
\end{table}

\begin{table}
	
	\caption{\label{mevrealshuf1} OzaBagADWIN - Shuffled UCI streams
	} 
	\scriptsize
	\centering
	\makebox[0.6\textwidth]{
		\begin{tabular}{p{6cm}?P{1.3cm}?P{1.3cm}?P{1.3cm}?P{1.3cm}}
			\toprule
			\multicolumn{1}{c?}{\textbf{Streams}} &  \multicolumn{2}{m{2.9cm}}{\textbf{VFDT Base}} &
			\multicolumn{2}{m{2.9cm}}{\textbf{EFDT Base}}\\
				\multicolumn{1}{c?}{} & Error & Variance & Error & Variance
				\\
			\midrule
			airlines & \textbf{0.3794} & 0.00025 & 0.38762 & 0.00026 \\
aws---price-discretized & 0.14701 & 0.00013 & \textbf{0.14066} & 0.00012 \\
chess & 0.67127 & 0.00022 & \textbf{0.59529} & 0.00029 \\
covertype & 0.27194 & 0.00021 & \textbf{0.25837} & 2e-04 \\
covpokelec & \textbf{0.3107} & 0.00037 & 0.37881 & 0.00065 \\
fonts & \textit{\textbf{0.00075}} & 0 & \textit{\textbf{0.00075}} & 0 \\
hhar & 0.05104 & 6e-05 & \textbf{0.03142} & 3e-05 \\
kdd & 0.00096 & 0 & \textbf{0.00079} & 0 \\
localization & 0.35122 & 0.00023 & \textbf{0.32013} & 0.00061 \\
miniboone & 0.10225 & 0.00011 & \textbf{0.10132} & 0.00011 \\
nbaiot & 0.33052 & 0.0013 & \textbf{0.0037} & 3e-05 \\
nswelec & 0.23162 & 0.00019 & \textbf{0.23045} & 0.00019 \\
pamap2 & 0.11963 & 0.00012 & \textbf{0.10477} & 0.00014 \\
poker & \textbf{0.26353} & 0.00024 & 0.28535 & 0.00047 \\
pucrio & 0.10893 & 0.00019 & \textbf{0.0449} & 9e-05 \\
sensor---home-activity & 0.0638 & 9e-05 & \textbf{0.04238} & 0.00019 \\
sensor---CO-discretized & 0.18565 & 0.00024 & \textbf{0.18344} & 0.00028 \\
skin & 0.01661 & 2e-05 & \textbf{0.00973} & 1e-05 \\
tnelec & \textit{\textbf{0.00678}} & 0 & \textit{\textbf{0.00678}} & 0 \\
wisdm & 0.17588 & 0.00018 & \textbf{0.11891} & 0.00012 \\
\midrule
\textbf{Unique Wins} & \multicolumn{2}{l?}{\textbf{3}} & \multicolumn{2}{l}{\textbf{15}} \\
\midrule\multicolumn{5}{l}{A \textbf{bold} error value indicates higher accuracy, and \textit{\textbf{bold italics}} indicate a tie.}\\
\multicolumn{5}{l}{\textbf{One-tailed binomial test statistics}:  \textbf{p-value: 0.00377};   \textbf{Confidence Interval:  0.62332 --- 1} }\\

			\bottomrule
		\end{tabular}}
\end{table}

\begin{table}[!ht]
	
	\caption{\label{mevrealshuf2} LevBagNoADWIN - Shuffled UCI streams
	} 
	\scriptsize
	\centering
	\makebox[0.6\textwidth]{
			\begin{tabular}{p{6cm}?P{1.3cm}?P{1.3cm}?P{1.3cm}?P{1.3cm}}
				\toprule
				\multicolumn{1}{c?}{\textbf{Streams}} &  \multicolumn{2}{m{2.9cm}}{\textbf{VFDT Base}} &
				\multicolumn{2}{m{2.9cm}}{\textbf{EFDT Base}}\\
				\multicolumn{1}{c?}{} & Error & Variance & Error & Variance
				\\
				\midrule
				airlines & \textbf{0.35897} & 0.00023 & 0.36134 & 0.00024 \\
aws---price-discretized & 0.13032 & 0.00011 & \textbf{0.12266} & 0.00011 \\
chess & 0.61357 & 0.00032 & \textbf{0.60015} & 0.00037 \\
covertype & 0.22738 & 0.00019 & \textbf{0.22208} & 0.00019 \\
covpokelec & \textbf{0.19074} & 0.00034 & 0.28548 & 0.00091 \\
fonts & \textit{\textbf{0.00061}} & 0 & \textit{\textbf{0.00061}} & 0 \\
hhar & 0.02496 & 3e-05 & \textbf{0.01873} & 2e-05 \\
kdd & 0.00064 & 0 & \textbf{0.00054} & 0 \\
localization & 0.33575 & 0.00047 & \textbf{0.30653} & 0.00072 \\
miniboone & \textbf{0.08894} & 8e-05 & 0.09681 & 0.00011 \\
nbaiot & 0.04423 & 0.00346 & \textbf{0.03933} & 0.00354 \\
nswelec & \textbf{0.22097} & 2e-04 & 0.22363 & 2e-04 \\
pamap2 & \textbf{0.0833} & 1e-04 & 0.10028 & 0.00015 \\
poker & \textbf{0.16567} & 0.00042 & 0.21305 & 0.00218 \\
pucrio & 0.0453 & 0.00012 & \textbf{0.04027} & 0.00024 \\
sensor---home-activity & \textbf{0.01491} & 2e-05 & 0.01644 & 6e-05 \\
sensor---CO-discretized & 0.1251 & 0.00012 & \textbf{0.12438} & 2e-04 \\
skin & 0.00529 & 1e-05 & \textbf{0.00394} & 0 \\
tnelec & \textit{\textbf{0.00356}} & 0 & \textit{\textbf{0.00356}} & 0 \\
wisdm & 0.13369 & 0.00013 & \textbf{0.11135} & 0.00013 \\
\midrule
\textbf{Unique Wins} & \multicolumn{2}{l?}{\textbf{7}} & \multicolumn{2}{l}{\textbf{11}} \\
\midrule\multicolumn{5}{l}{A \textbf{bold} error value indicates higher accuracy, and \textit{\textbf{bold italics}} indicate a tie.}\\
\multicolumn{5}{l}{\textbf{One-tailed binomial test statistics}:  \textbf{p-value: 0.24034};   \textbf{Confidence Interval:  0.39216 --- 1} }\\

				\bottomrule
		\end{tabular}}
\end{table}

\begin{table}[!ht]
	
	\caption{\label{mevrealshuf3} LevBag - Shuffled UCI streams
	} 
	\scriptsize
	\centering
	\makebox[0.6\textwidth]{
			\begin{tabular}{p{6cm}?P{1.3cm}?P{1.3cm}?P{1.3cm}?P{1.3cm}}
				\toprule
				\multicolumn{1}{c?}{\textbf{Streams}} &  \multicolumn{2}{m{2.9cm}}{\textbf{VFDT Base}} &
				\multicolumn{2}{m{2.9cm}}{\textbf{EFDT Base}}\\
				\multicolumn{1}{c?}{} & Error & Variance & Error & Variance
				\\
				\midrule
				airlines & \textbf{0.41108} & 0.00032 & 0.42212 & 0.00032 \\
aws---price-discretized & 0.1311 & 0.00011 & \textbf{0.12379} & 0.00011 \\
chess & \textbf{0.62328} & 0.00071 & 0.63585 & 0.00083 \\
covertype & 0.22735 & 0.00019 & \textbf{0.22286} & 2e-04 \\
covpokelec & \textbf{0.18993} & 0.00023 & 0.36941 & 0.00084 \\
fonts & \textit{\textbf{6e-04}} & 0 & \textit{\textbf{6e-04}} & 0 \\
hhar & 0.02357 & 3e-05 & \textbf{0.01815} & 3e-05 \\
kdd & 0.00063 & 0 & \textbf{0.00054} & 0 \\
localization & \textbf{0.38614} & 0.00069 & 0.44004 & 0.00041 \\
miniboone & \textbf{0.08798} & 8e-05 & 0.09766 & 0.00011 \\
nbaiot & 0.10372 & 0.00349 & \textbf{0.00304} & 3e-05 \\
nswelec & \textbf{0.22097} & 2e-04 & 0.22377 & 2e-04 \\
pamap2 & \textbf{0.33327} & 0.00182 & 0.36266 & 0.00155 \\
poker & \textbf{0.16436} & 4e-04 & 0.25652 & 0.00141 \\
pucrio & 0.04534 & 0.00011 & \textbf{0.04214} & 0.00014 \\
sensor---home-activity & \textbf{0.01491} & 2e-05 & 0.02001 & 0.00011 \\
sensor---CO-discretized & \textbf{0.12546} & 0.00012 & 0.13818 & 0.00021 \\
skin & 0.00529 & 1e-05 & \textbf{0.00399} & 0 \\
tnelec & \textit{\textbf{0.00356}} & 0 & \textit{\textbf{0.00356}} & 0 \\
wisdm & 0.13278 & 0.00013 & \textbf{0.11152} & 0.00013 \\
\midrule
\textbf{Unique Wins} & \multicolumn{2}{l?}{\textbf{10}} & \multicolumn{2}{l}{\textbf{8}} \\
\midrule\multicolumn{5}{l}{A \textbf{bold} error value indicates higher accuracy, and \textit{\textbf{bold italics}} indicate a tie.}\\
\multicolumn{5}{l}{\textbf{One-tailed binomial test statistics}:  \textbf{p-value: 0.75966};   \textbf{Confidence Interval:  0.24396 --- 1} }\\

				\bottomrule
		\end{tabular}}
	
\end{table}

\begin{table}[!ht]
	
	\caption{\label{mevrealshuf4} Adaptive Random Forest - Shuffled UCI streams
	} 
	\scriptsize
	\centering
	\makebox[0.6\textwidth]{
			\begin{tabular}{p{6cm}?P{1.3cm}?P{1.3cm}?P{1.3cm}?P{1.3cm}}
				\toprule
				\multicolumn{1}{c?}{\textbf{Streams}} &  \multicolumn{2}{P{2.6cm}}{\textbf{VFDT Base}} &
				\multicolumn{2}{P{2.6cm}}{\textbf{EFDT Base}}\\
				\multicolumn{1}{c?}{} & Error & Variance & Error & Variance 
				\\
				\midrule
				airlines & 0.40058 & 0.00033 & \textbf{0.39267} & 0.00028 \\
aws---price-discretized & \textbf{0.18202} & 0.00021 & 0.19533 & 0.00027 \\
chess & 0.70999 & 0.00062 & \textbf{0.63728} & 0.00046 \\
covertype & 0.2876 & 0.00024 & \textbf{0.2796} & 0.00025 \\
covpokelec & 0.48633 & 0.00075 & \textbf{0.44633} & 0.00054 \\
fonts & 0.80344 & 5e-04 & \textbf{0.68368} & 0.00081 \\
hhar & 0.82982 & 0.00054 & \textbf{0.33478} & 0.00069 \\
kdd & \textbf{0.00191} & 0 & 0.00284 & 0 \\
localization & 0.43867 & 0.00031 & \textbf{0.43129} & 0.00056 \\
miniboone & 0.10704 & 0.00014 & \textbf{0.10339} & 0.00011 \\
nbaiot & 0.28421 & 0.00232 & \textbf{0.11822} & 0.00154 \\
nswelec & \textbf{0.22344} & 0.00021 & 0.2238 & 2e-04 \\
pamap2 & \textbf{0.21005} & 0.00036 & 0.27404 & 0.00324 \\
poker & 0.37042 & 0.00031 & \textbf{0.35486} & 0.00029 \\
pucrio & 0.16029 & 0.00033 & \textbf{0.0893} & 0.00017 \\
sensor---home-activity & 0.32884 & 0.00049 & \textbf{0.28902} & 0.00044 \\
sensor---CO-discretized & 0.39529 & 5e-04 & \textbf{0.32805} & 0.00044 \\
skin & 0.19835 & 0.00033 & \textbf{0.04533} & 0.00023 \\
tnelec & 0.35672 & 0.01894 & \textbf{0.28284} & 0.00168 \\
wisdm & 0.22926 & 0.00045 & \textbf{0.1882} & 0.00032 \\
\midrule
\textbf{Unique Wins} & \multicolumn{2}{l?}{\textbf{4}} & \multicolumn{2}{l}{\textbf{16}} \\
\midrule\multicolumn{5}{l}{A \textbf{bold} error value indicates higher accuracy, and \textit{\textbf{bold italics}} indicate a tie.}\\
\multicolumn{5}{l}{\textbf{One-tailed binomial test statistics}:  \textbf{p-value: 0.00591};   \textbf{Confidence Interval:  0.59897 --- 1} }\\

				\bottomrule
		\end{tabular}}
\end{table}

	
\begin{table}[!ht]
	
	\caption{\label{mevrealshuf5} OzaBoost - Shuffled UCI streams
	} 
	\scriptsize
	\centering
	\makebox[0.6\textwidth]{
			\begin{tabular}{p{6cm}?P{1.3cm}?P{1.3cm}?P{1.3cm}?P{1.3cm}}
				\toprule
				\multicolumn{1}{c?}{\textbf{Streams}} &  \multicolumn{2}{m{2.9cm}}{\textbf{VFDT Base}} &
				\multicolumn{2}{m{2.9cm}}{\textbf{EFDT Base}}\\
				\multicolumn{1}{c?}{} & Error & Variance & Error & Variance
				\\
				\midrule
				airlines & \textbf{0.38601} & 0.00024 & 0.39146 & 0.00025 \\
aws---price-discretized & 0.14527 & 0.00012 & \textbf{0.13761} & 0.00012 \\
chess & \textit{\textbf{0.90007}} & 1e-04 & \textit{\textbf{0.90007}} & 1e-04 \\
covertype & 0.26591 & 0.00022 & \textbf{0.25115} & 0.00021 \\
covpokelec & \textbf{0.29062} & 0.00053 & 0.34335 & 0.0013 \\
fonts & \textit{\textbf{0.00107}} & 0 & \textit{\textbf{0.00107}} & 0 \\
hhar & 0.02277 & 7e-05 & \textbf{0.00953} & 1e-05 \\
kdd & \textbf{0.00034} & 0 & 0.00035 & 0 \\
localization & 0.33213 & 5e-04 & \textbf{0.31382} & 0.00072 \\
miniboone & \textbf{0.09859} & 0.00021 & 0.10123 & 0.00012 \\
nbaiot & 0.02137 & 0.00022 & \textbf{0.00879} & 0.00013 \\
nswelec & \textbf{0.23942} & 0.00029 & 0.24197 & 0.00031 \\
pamap2 & \textbf{0.08673} & 0.00015 & 0.08794 & 0.00014 \\
poker & \textbf{0.23568} & 0.00032 & 0.24227 & 0.00075 \\
pucrio & 0.07653 & 3e-04 & \textbf{0.02393} & 4e-05 \\
sensor---home-activity & 0.04266 & 7e-05 & \textbf{0.02925} & 6e-05 \\
sensor---CO-discretized & 0.15979 & 0.00071 & \textbf{0.13586} & 0.00056 \\
skin & 0.00485 & 1e-05 & \textbf{0.0033} & 1e-05 \\
tnelec & \textit{\textbf{0.0065}} & 2e-05 & \textit{\textbf{0.0065}} & 2e-05 \\
wisdm & 0.17925 & 0.00022 & \textbf{0.1267} & 0.00014 \\
\midrule
\textbf{Unique Wins} & \multicolumn{2}{l?}{\textbf{7}} & \multicolumn{2}{l}{\textbf{10}} \\
\midrule\multicolumn{5}{l}{A \textbf{bold} error value indicates higher accuracy, and \textit{\textbf{bold italics}} indicate a tie.}\\
\multicolumn{5}{l}{\textbf{One-tailed binomial test statistics}:  \textbf{p-value: 0.31453};   \textbf{Confidence Interval:  0.36401 --- 1} }\\

				\bottomrule
		\end{tabular}}
\end{table}

\begin{table}[!ht]
	
	\caption{\label{mevrealshuf6} OzaBoostADWIN - Shuffled UCI streams
	} 
	\scriptsize
	\centering
	\makebox[0.6\textwidth]{
			\begin{tabular}{p{6cm}?P{1.3cm}?P{1.3cm}?P{1.3cm}?P{1.3cm}}
				\toprule
				\multicolumn{1}{c?}{\textbf{Streams}} &  \multicolumn{2}{m{2.9cm}}{\textbf{VFDT Base}} &
				\multicolumn{2}{m{2.9cm}}{\textbf{EFDT Base}}\\
				\multicolumn{1}{c?}{} & Error & Variance & Error & Variance
				\\
				\midrule
				airlines & \textbf{0.43251} & 0.00028 & 0.4347 & 0.00029 \\
aws---price-discretized & 0.14084 & 0.00019 & \textbf{0.1388} & 0.00013 \\
chess & \textbf{0.89967} & 0.00011 & 0.89985 & 1e-04 \\
covertype & 0.26289 & 0.00035 & \textbf{0.26169} & 0.00034 \\
covpokelec & \textbf{0.34818} & 0.00481 & 0.40584 & 0.00587 \\
fonts & \textit{\textbf{0.00107}} & 0 & \textit{\textbf{0.00107}} & 0 \\
hhar & 0.02243 & 0.00012 & \textbf{0.01147} & 2e-05 \\
kdd & 0.00072 & 5e-05 & \textbf{0.00061} & 0 \\
localization & 0.33738 & 0.0014 & \textbf{0.31842} & 0.00141 \\
miniboone & \textbf{0.10724} & 0.00014 & 0.12055 & 2e-04 \\
nbaiot & 0.02992 & 0.00104 & \textbf{0.00998} & 0.00033 \\
nswelec & \textbf{0.26887} & 0.00045 & 0.27118 & 0.00041 \\
pamap2 & \textbf{0.06545} & 0.00018 & 0.09245 & 0.00027 \\
poker & \textbf{0.27171} & 0.00135 & 0.305 & 0.0026 \\
pucrio & 0.0417 & 5e-04 & \textbf{0.02821} & 5e-05 \\
sensor---home-activity & 0.03694 & 0.00039 & \textbf{0.03507} & 0.00016 \\
sensor---CO-discretized & 0.16919 & 0.00097 & \textbf{0.1599} & 0.00142 \\
skin & 0.00362 & 1e-05 & \textbf{0.00329} & 1e-05 \\
tnelec & \textit{\textbf{0.00787}} & 0.00026 & \textit{\textbf{0.00787}} & 0.00026 \\
wisdm & 0.14083 & 0.00052 & \textbf{0.12696} & 0.00014 \\
\midrule
\textbf{Unique Wins} & \multicolumn{2}{l?}{\textbf{7}} & \multicolumn{2}{l}{\textbf{11}} \\
\midrule\multicolumn{5}{l}{A \textbf{bold} error value indicates higher accuracy, and \textit{\textbf{bold italics}} indicate a tie.}\\
\multicolumn{5}{l}{\textbf{One-tailed binomial test statistics}:  \textbf{p-value: 0.24034};   \textbf{Confidence Interval:  0.39216 --- 1} }\\

				\bottomrule
		\end{tabular}}
\end{table}

\begin{table}[!ht]
	
	\caption{\label{mevrealshuf7} ADOB - Shuffled UCI streams
	} 
	\scriptsize
	\centering
	\makebox[0.6\textwidth]{
			\begin{tabular}{p{6cm}?P{1.3cm}?P{1.3cm}?P{1.3cm}?P{1.3cm}}
				\toprule
				\multicolumn{1}{c?}{\textbf{Streams}} &  \multicolumn{2}{m{2.9cm}}{\textbf{VFDT Base}} &
				\multicolumn{2}{m{2.9cm}}{\textbf{EFDT Base}}\\
				\multicolumn{1}{c?}{} & Error & Variance & Error & Variance
				\\
				\midrule
				airlines & \textbf{0.38375} & 0.00025 & 0.38679 & 0.00025 \\
aws---price-discretized & 0.14249 & 0.00012 & \textbf{0.13355} & 0.00012 \\
chess & 0.89206 & 0.00177 & \textbf{0.75812} & 0.02005 \\
covertype & 0.2607 & 0.00021 & \textbf{0.24908} & 2e-04 \\
covpokelec & \textbf{0.27884} & 0.00043 & 0.3403 & 0.00113 \\
fonts & \textit{\textbf{0.00102}} & 1e-05 & \textit{\textbf{0.00102}} & 1e-05 \\
hhar & 0.02323 & 8e-05 & \textbf{0.00933} & 1e-05 \\
kdd & \textbf{0.00032} & 0 & 0.00033 & 0 \\
localization & 0.33964 & 0.00042 & \textbf{0.32868} & 0.00093 \\
miniboone & 0.10164 & 0.00039 & \textbf{0.10128} & 0.00012 \\
nbaiot & 0.01748 & 0.00064 & \textbf{0.0053} & 7e-05 \\
nswelec & \textbf{0.2455} & 0.00033 & 0.2483 & 0.00035 \\
pamap2 & 0.08814 & 0.00015 & \textbf{0.08401} & 0.00016 \\
poker & \textbf{0.22355} & 0.00044 & 0.2413 & 0.00089 \\
pucrio & 0.06842 & 0.00025 & \textbf{0.02511} & 5e-05 \\
sensor---home-activity & 0.04036 & 9e-05 & \textbf{0.02613} & 6e-05 \\
sensor---CO-discretized & 0.14869 & 0.00075 & \textbf{0.12618} & 0.00043 \\
skin & 0.00552 & 2e-05 & \textbf{0.00384} & 1e-05 \\
tnelec & \textit{\textbf{0.00646}} & 2e-05 & \textit{\textbf{0.00646}} & 2e-05 \\
wisdm & 0.17089 & 2e-04 & \textbf{0.12382} & 0.00013 \\
\midrule
\textbf{Unique Wins} & \multicolumn{2}{l?}{\textbf{5}} & \multicolumn{2}{l}{\textbf{13}} \\
\midrule\multicolumn{5}{l}{A \textbf{bold} error value indicates higher accuracy, and \textit{\textbf{bold italics}} indicate a tie.}\\
\multicolumn{5}{l}{\textbf{One-tailed binomial test statistics}:  \textbf{p-value: 0.04813};   \textbf{Confidence Interval:  0.50217 --- 1} }\\

				\bottomrule
		\end{tabular}}
\end{table}

\begin{table}[!ht]
	
	\caption{\label{mevrealshuf8} BOLE - Shuffled UCI streams
	} 
	\scriptsize
	\centering
	\makebox[0.6\textwidth]{
			\begin{tabular}{p{6cm}?P{1.3cm}?P{1.3cm}?P{1.3cm}?P{1.3cm}}
				\toprule
				\multicolumn{1}{c?}{\textbf{Streams}} &  \multicolumn{2}{m{2.9cm}}{\textbf{VFDT Base}} &
				\multicolumn{2}{m{2.9cm}}{\textbf{EFDT Base}}\\
				\multicolumn{1}{c?}{} & Error & Variance & Error & Variance
				\\
				\midrule
				airlines & \textbf{0.38375} & 0.00025 & 0.3868 & 0.00025 \\
aws---price-discretized & 0.14249 & 0.00012 & \textbf{0.13355} & 0.00012 \\
chess & 0.83787 & 0.00973 & \textbf{0.60847} & 0.00046 \\
covertype & 0.26065 & 0.00021 & \textbf{0.24906} & 2e-04 \\
covpokelec & \textbf{0.27706} & 0.00041 & 0.33397 & 0.00093 \\
fonts & \textit{\textbf{0.00087}} & 0 & \textit{\textbf{0.00087}} & 0 \\
hhar & 0.02299 & 4e-05 & \textbf{0.00933} & 1e-05 \\
kdd & \textbf{0.00032} & 0 & 0.00033 & 0 \\
localization & 0.33935 & 0.00041 & \textbf{0.32839} & 0.00092 \\
miniboone & 0.10164 & 0.00039 & \textbf{0.10127} & 0.00012 \\
nbaiot & 0.01617 & 2e-04 & \textbf{0.0053} & 7e-05 \\
nswelec & \textbf{0.2455} & 0.00033 & 0.2483 & 0.00035 \\
pamap2 & 0.08814 & 0.00015 & \textbf{0.08401} & 0.00016 \\
poker & \textbf{0.22348} & 0.00044 & 0.24122 & 0.00088 \\
pucrio & 0.0684 & 0.00025 & \textbf{0.02508} & 5e-05 \\
sensor---home-activity & 0.04023 & 6e-05 & \textbf{0.02611} & 6e-05 \\
sensor---CO-discretized & 0.14675 & 0.00022 & \textbf{0.12573} & 0.00029 \\
skin & 0.00551 & 2e-05 & \textbf{0.00382} & 1e-05 \\
tnelec & \textit{\textbf{0.00583}} & 0 & \textit{\textbf{0.00583}} & 0 \\
wisdm & 0.17085 & 2e-04 & \textbf{0.12378} & 0.00013 \\
\midrule
\textbf{Unique Wins} & \multicolumn{2}{l?}{\textbf{5}} & \multicolumn{2}{l}{\textbf{13}} \\
\midrule\multicolumn{5}{l}{A \textbf{bold} error value indicates higher accuracy, and \textit{\textbf{bold italics}} indicate a tie.}\\
\multicolumn{5}{l}{\textbf{One-tailed binomial test statistics}:  \textbf{p-value: 0.04813};   \textbf{Confidence Interval:  0.50217 --- 1} }\\

				\bottomrule
		\end{tabular}}
\end{table}

\begin{table}[!ht]
	
	\caption{\label{mevrealshuf9} OnlineSmoothBoost - Shuffled UCI streams
	} 
	\scriptsize
	\centering
	\makebox[0.6\textwidth]{
		\begin{tabular}{p{6cm}?P{1.3cm}?P{1.3cm}?P{1.3cm}?P{1.3cm}}
			\toprule
			\multicolumn{1}{c?}{\textbf{Streams}} &  \multicolumn{2}{m{2.9cm}}{\textbf{VFDT Base}} &
			\multicolumn{2}{m{2.9cm}}{\textbf{EFDT Base}}\\
				\multicolumn{1}{c?}{} & Error & Variance & Error & Variance
				\\
			\midrule
			airlines & \textbf{0.35598} & 0.00023 & 0.35792 & 0.00023 \\
aws---price-discretized & 0.14553 & 0.00012 & \textbf{0.13925} & 0.00012 \\
chess & 0.6674 & 0.00023 & \textbf{0.58559} & 0.00035 \\
covertype & 0.27164 & 0.00021 & \textbf{0.26133} & 0.00021 \\
covpokelec & \textbf{0.3003} & 3e-04 & 0.33391 & 0.00073 \\
fonts & \textit{\textbf{0.00068}} & 0 & \textit{\textbf{0.00068}} & 0 \\
hhar & 0.05521 & 8e-05 & \textbf{0.0363} & 4e-05 \\
kdd & 0.00106 & 0 & \textbf{0.00082} & 0 \\
localization & 0.34715 & 0.00032 & \textbf{0.32556} & 9e-04 \\
miniboone & 0.10497 & 0.00011 & \textbf{0.10297} & 0.00015 \\
nbaiot & 0.03079 & 0.00032 & \textbf{0.01775} & 0.00174 \\
nswelec & 0.23471 & 0.00021 & \textbf{0.23173} & 0.00021 \\
pamap2 & 0.12848 & 0.00017 & \textbf{0.10889} & 0.00019 \\
poker & \textbf{0.26615} & 0.00027 & 0.2724 & 0.00084 \\
pucrio & 0.12499 & 0.00039 & \textbf{0.05627} & 0.00045 \\
sensor---home-activity & 0.07077 & 0.00012 & \textbf{0.03824} & 0.00012 \\
sensor---CO-discretized & 0.18328 & 0.00023 & \textbf{0.16407} & 0.00028 \\
skin & 0.01677 & 3e-05 & \textbf{0.01178} & 2e-05 \\
tnelec & \textit{\textbf{0.00588}} & 1e-05 & \textit{\textbf{0.00588}} & 1e-05 \\
wisdm & 0.18723 & 0.00022 & \textbf{0.13099} & 0.00018 \\
\midrule
\textbf{Unique Wins} & \multicolumn{2}{l?}{\textbf{3}} & \multicolumn{2}{l}{\textbf{15}} \\
\midrule\multicolumn{5}{l}{A \textbf{bold} error value indicates higher accuracy, and \textit{\textbf{bold italics}} indicate a tie.}\\
\multicolumn{5}{l}{\textbf{One-tailed binomial test statistics}:  \textbf{p-value: 0.00377};   \textbf{Confidence Interval:  0.62332 --- 1} }\\

			\bottomrule
		\end{tabular}}
\end{table}

\begin{table}[!ht]
	
	\caption{\label{mevrealshuf10} Plain single learners (no ensemble) - Shuffled UCI streams
	} 
	\scriptsize
	\centering
	\makebox[0.6\textwidth]{
		\begin{tabular}{p{6cm}?P{1.3cm}?P{1.3cm}?P{1.3cm}?P{1.3cm}}
			\toprule
			\multicolumn{1}{c?}{\textbf{Streams}} &  \multicolumn{2}{m{2.9cm}}{\textbf{VFDT Base}} &
			\multicolumn{2}{m{2.9cm}}{\textbf{EFDT Base}}\\
				\multicolumn{1}{c?}{} & Error & Variance & Error & Variance
				\\
			\midrule
			airlines & \textbf{0.35926} & 0.00023 & 0.36099 & 0.00023 \\
aws---price-discretized & 0.14715 & 0.00013 & \textbf{0.14139} & 0.00012 \\
chess & 0.67174 & 0.00022 & \textbf{0.60706} & 0.00058 \\
covertype & 0.28618 & 0.00023 & \textbf{0.27648} & 0.00022 \\
covpokelec & \textbf{0.32762} & 0.00044 & 0.37454 & 0.00207 \\
fonts & \textit{\textbf{0.00068}} & 0 & \textit{\textbf{0.00068}} & 0 \\
hhar & 0.07311 & 8e-05 & \textbf{0.05113} & 6e-05 \\
kdd & 0.00114 & 0 & \textbf{0.00094} & 0 \\
localization & 0.35944 & 0.00025 & \textbf{0.33857} & 0.00128 \\
miniboone & \textbf{0.1163} & 0.00015 & 0.11875 & 0.00022 \\
nbaiot & 0.0381 & 0.00165 & \textbf{0.02908} & 0.00275 \\
nswelec & 0.24141 & 0.00027 & \textbf{0.24057} & 0.00023 \\
pamap2 & 0.15764 & 0.00032 & \textbf{0.15672} & 3e-04 \\
poker & \textbf{0.28636} & 0.00032 & 0.30909 & 0.00142 \\
pucrio & 0.13639 & 5e-04 & \textbf{0.08329} & 0.00084 \\
sensor---home-activity & 0.11511 & 0.00022 & \textbf{0.0955} & 0.00187 \\
sensor---CO-discretized & \textbf{0.23193} & 4e-04 & 0.2432 & 0.00145 \\
skin & 0.01908 & 3e-05 & \textbf{0.01359} & 3e-05 \\
tnelec & \textit{\textbf{0.00586}} & 1e-05 & \textit{\textbf{0.00586}} & 1e-05 \\
wisdm & 0.1888 & 0.00021 & \textbf{0.13637} & 0.00019 \\
\midrule
\textbf{Unique Wins} & \multicolumn{2}{l?}{\textbf{5}} & \multicolumn{2}{l}{\textbf{13}} \\
\midrule\multicolumn{5}{l}{A \textbf{bold} error value indicates higher accuracy, and \textit{\textbf{bold italics}} indicate a tie.}\\
\multicolumn{5}{l}{\textbf{One-tailed binomial test statistics}:  \textbf{p-value: 0.04813};   \textbf{Confidence Interval:  0.50217 --- 1} }\\

			\bottomrule
		\end{tabular}}
\end{table}

	The pattern of comparative prequential accuracy performance is roughly the same for three of the bagging approaches, but not for the Leveraged Bagging approaches. OzaBag (Table \ref{mevrealshuf0}), OzaBagADWIN (Table \ref{mevrealshuf1}) and Adaptive Random Forest (Table \ref{mevrealshuf4}) all favor EFDT with p-values of 0.00377, 0.00377, and 0.00591 respectively. However, in the absence of residual concept drift, it appears that EFDT based Leveraging Bagging without ADWIN (Table \ref{mevrealshuf2}, p-value 0.24034) does not attain significance, and Leveraging Bagging (Table \ref{mevrealshuf3}, p-value 0.75966) also loses significance (VFDT outperforms slightly, 10 wins to 8 and 2 draws). Change detection with Leveraging Bagging appears to make the effect more severe, but changing the choice of Poisson value alone (to 6) as compared to OzaBag negates the advantage of EFDT when the stream is highly uniform. This erosion of advantage for EFDT based Leveraging Bagging when supplied highly uniform streams merits further investigation.

	Among the boosting approaches, we note the failure of OzaBoost and OzaBoostADWIN to reach significance (Tables \ref{mevrealshuf5}, \ref{mevrealshuf6}, p-values 0.31453 and 0.24034 respectively), and the success of ADOB, BOLE and OnlineSmoothBoost (Tables \ref{mevrealshuf7}, \ref{mevrealshuf8}, and \ref{mevrealshuf9}, p-values 0.04813, 0.04813, and 0.00377 respectively) well within a 0.05 significance level. Clearly, improving strategies for weighting misclassified examples is of particular interest for future work on boosting in general.
	
	Plain EFDT outperforms plain VFDT on 13 streams (Table \ref{mevrealshuf10}), with VFDT registering lower prequential accuracy on 5 and drawing on 2.
	
	Our ranked p-values are 0.00377, 0.00377, 0.00377, 0.00591, 0.04813, 0.04813, 0.24034, 0.24034, 0.31453, 0.75966. The first four alternate hypotheses would be rejected using a Holm-Bonferroni multiple testing correction, resulting in findings of significant improvements in the number of stationary data streams for which the ensemble technique is more accurate using EFDT than VFDT for OzaBag, OzaBagAdwin,  Adaptive Random Forest and OnlineSmoothBoost.

	\FloatBarrier
	
	\subsection{Synthetic Streams}
	\label{sec:synthetic}

	Tables \ref{mevsyn0} through \ref{mevsyn10} compare EFDT and VFDT based ensembles on a large number of parameterized synthetic concept drift streams found in the literature that are described in detail in Section~\ref{expsetup}. 
	
	\begin{table}
	
	\caption{\label{mevsyn0} OzaBag - Synthetic streams with concept drift
	} 
	\scriptsize
	\centering
	\makebox[0.8\textwidth]{
			\begin{tabular}{p{6cm}?P{1.3cm}?P{1.3cm}?P{1.3cm}?P{1.3cm}}
				\toprule
				\multicolumn{1}{c?}{\textbf{Streams}} &  \multicolumn{2}{m{2.9cm}}{\textbf{VFDT Base}} &
				\multicolumn{2}{m{2.9cm}}{\textbf{EFDT Base}}\\
				\multicolumn{1}{c?}{} & Error & Variance & Error & Variance
				\\
				\midrule
				recurrent---agrawal & 0.1968 & 0.00031 & \textbf{0.18373} & 0.00031 \\
recurrent---led & \textbf{0.32315} & 0.00027 & 0.32699 & 0.00026 \\
recurrent---randomtree & 0.21247 & 0.0022 & \textbf{0.20024} & 0.00299 \\
recurrent---sea & 0.14635 & 0.00014 & \textbf{0.13948} & 0.00013 \\
recurrent---stagger & \textbf{0.17477} & 0.00123 & 0.17949 & 0.00078 \\
recurrent---waveform & 0.16931 & 0.0015 & \textbf{0.16033} & 0.0012 \\
hyperplane---1 & 0.10901 & 0.00019 & \textbf{0.10626} & 0.00014 \\
hyperplane---2 & 0.15844 & 0.00119 & \textbf{0.12861} & 0.00056 \\
hyperplane---3 & \textbf{0.10194} & 0.00012 & 0.10271 & 0.00011 \\
hyperplane---4 & 0.15448 & 0.00349 & \textbf{0.12442} & 0.00131 \\
rbf---drift-1 & 0.07945 & 0.00032 & \textbf{0.07194} & 0.00026 \\
rbf---drift-2 & 0.18388 & 0.00088 & \textbf{0.16055} & 0.00094 \\
rbf---drift-3 & 0.11286 & 0.00063 & \textbf{0.10982} & 6e-04 \\
rbf---drift-4 & 0.36223 & 0.00134 & \textbf{0.35289} & 0.00127 \\
recurrent---abrupt---222 & 0.3431 & 0.02363 & \textbf{0.33746} & 0.02201 \\
recurrent---abrupt---322 & 0.37099 & 0.01543 & \textbf{0.35744} & 0.03263 \\
recurrent---abrupt---332 & 0.3459 & 0.08753 & \textbf{0.32838} & 0.10325 \\
recurrent---abrupt---333 & 0.36143 & 0.01448 & \textbf{0.35037} & 0.04007 \\
recurrent---abrupt---334 & 0.39642 & 0.00214 & \textbf{0.3698} & 0.05898 \\
recurrent---abrupt---335 & 0.39614 & 0.00314 & \textbf{0.38665} & 0.04395 \\
recurrent---abrupt---422 & \textbf{0.34057} & 0.02736 & 0.34205 & 0.04047 \\
recurrent---abrupt---444 & 0.40099 & 0.00751 & \textbf{0.36757} & 0.01922 \\
recurrent---abrupt---522 & \textbf{0.33857} & 0.0361 & 0.34167 & 0.07319 \\
recurrent---abrupt---555 & 0.40625 & 0.00469 & \textbf{0.35437} & 0.00596 \\
\midrule
\textbf{Unique Wins} & \multicolumn{2}{l?}{\textbf{5}} & \multicolumn{2}{l}{\textbf{19}} \\
\midrule\multicolumn{5}{l}{A \textbf{bold} error value indicates higher accuracy, and \textit{\textbf{bold italics}} indicate a tie.}\\
\multicolumn{5}{l}{\textbf{One-tailed binomial test statistics}:  \textbf{p-value: 0.00331};   \textbf{Confidence Interval:  0.61086 --- 1} }\\

				\bottomrule
		\end{tabular}}
\end{table}

\begin{table}
	
	\caption{\label{mevsyn1} OzaBagADWIN - Synthetic streams with concept drift
	} 
	\scriptsize
	\centering
	\makebox[0.8\textwidth]{
		\begin{tabular}{p{6cm}?P{1.3cm}?P{1.3cm}?P{1.3cm}?P{1.3cm}}
			\toprule
			\multicolumn{1}{c?}{\textbf{Streams}} &  \multicolumn{2}{m{2.9cm}}{\textbf{VFDT Base}} &
			\multicolumn{2}{m{2.9cm}}{\textbf{EFDT Base}}\\
				\multicolumn{1}{c?}{} & Error & Variance & Error & Variance
				\\
			\midrule
			recurrent---agrawal & \textbf{0.11929} & 0.00097 & 0.12652 & 0.00042 \\
recurrent---led & \textbf{0.26175} & 2e-04 & 0.26285 & 2e-04 \\
recurrent---randomtree & 0.09511 & 0.00212 & \textbf{0.0846} & 0.00232 \\
recurrent---sea & 0.11452 & 0.00023 & \textbf{0.11262} & 0.00018 \\
recurrent---stagger & \textbf{0.00172} & 0.00014 & 0.00299 & 0.00065 \\
recurrent---waveform & 0.15625 & 0.00021 & \textbf{0.14762} & 0.00015 \\
hyperplane---1 & \textbf{0.10114} & 0.00012 & 0.10371 & 0.00012 \\
hyperplane---2 & \textbf{0.10998} & 0.00019 & 0.11387 & 0.00019 \\
hyperplane---3 & \textbf{0.09972} & 0.00011 & 0.10153 & 0.00011 \\
hyperplane---4 & \textbf{0.10515} & 0.00029 & 0.10753 & 0.00031 \\
rbf---drift-1 & 0.07738 & 0.00029 & \textbf{0.07201} & 0.00026 \\
rbf---drift-2 & 0.13282 & 0.00082 & \textbf{0.11136} & 0.00059 \\
rbf---drift-3 & 0.11143 & 0.00062 & \textbf{0.10947} & 6e-04 \\
rbf---drift-4 & 0.30834 & 0.00193 & \textbf{0.29876} & 0.00209 \\
recurrent---abrupt---222 & 0.12285 & 0.03384 & \textbf{0.11331} & 0.03473 \\
recurrent---abrupt---322 & 0.16658 & 0.02614 & \textbf{0.15517} & 0.02375 \\
recurrent---abrupt---332 & \textbf{0.11804} & 0.04584 & 0.12188 & 0.05251 \\
recurrent---abrupt---333 & 0.14935 & 0.01806 & \textbf{0.14827} & 0.02198 \\
recurrent---abrupt---334 & \textbf{0.1531} & 0.04044 & 0.17691 & 0.03843 \\
recurrent---abrupt---335 & 0.18661 & 0.0267 & \textbf{0.15904} & 0.05559 \\
recurrent---abrupt---422 & \textbf{0.12408} & 0.03756 & 0.1336 & 0.04196 \\
recurrent---abrupt---444 & 0.17194 & 0.01932 & \textbf{0.15267} & 0.04052 \\
recurrent---abrupt---522 & \textbf{0.11097} & 0.04375 & 0.11422 & 0.04112 \\
recurrent---abrupt---555 & 0.29009 & 0.00912 & \textbf{0.15781} & 0.01077 \\
\midrule
\textbf{Unique Wins} & \multicolumn{2}{l?}{\textbf{11}} & \multicolumn{2}{l}{\textbf{13}} \\
\midrule\multicolumn{5}{l}{A \textbf{bold} error value indicates higher accuracy, and \textit{\textbf{bold italics}} indicate a tie.}\\
\multicolumn{5}{l}{\textbf{One-tailed binomial test statistics}:  \textbf{p-value: 0.41941};   \textbf{Confidence Interval:  0.35756 --- 1} }\\

			\bottomrule
		\end{tabular}}
\end{table}

\begin{table}
	
	\caption{\label{mevsyn2} LevBagNoADWIN - Synthetic streams with concept drift
	} 
	\scriptsize
	\centering
	\makebox[0.8\textwidth]{
			\begin{tabular}{p{6cm}?P{1.3cm}?P{1.3cm}?P{1.3cm}?P{1.3cm}}
				\toprule
				\multicolumn{1}{c?}{\textbf{Streams}} &  \multicolumn{2}{m{2.9cm}}{\textbf{VFDT Base}} &
				\multicolumn{2}{m{2.9cm}}{\textbf{EFDT Base}}\\
				\multicolumn{1}{c?}{} & Error & Variance & Error & Variance
				\\
				\midrule
				recurrent---agrawal & 0.2001 & 0.00046 & \textbf{0.1739} & 0.00048 \\
recurrent---led & 0.32774 & 0.00024 & \textbf{0.31783} & 0.00033 \\
recurrent---randomtree & 0.19747 & 0.00163 & \textbf{0.18482} & 0.00199 \\
recurrent---sea & 0.14106 & 0.00013 & \textbf{0.13493} & 0.00015 \\
recurrent---stagger & \textbf{0.19141} & 0.00029 & 0.1926 & 0.00016 \\
recurrent---waveform & 0.16861 & 0.00146 & \textbf{0.16064} & 0.00085 \\
hyperplane---1 & 0.1141 & 0.00019 & \textbf{0.11174} & 0.00014 \\
hyperplane---2 & 0.16105 & 0.00108 & \textbf{0.1276} & 0.00042 \\
hyperplane---3 & \textbf{0.10639} & 0.00013 & 0.10955 & 0.00012 \\
hyperplane---4 & 0.15599 & 0.00347 & \textbf{0.12183} & 0.00098 \\
rbf---drift-1 & 0.06514 & 0.00017 & \textbf{0.06263} & 0.00017 \\
rbf---drift-2 & 0.15039 & 0.00048 & \textbf{0.11581} & 0.00051 \\
rbf---drift-3 & 0.09594 & 0.00043 & \textbf{0.09147} & 0.00038 \\
rbf---drift-4 & 0.30655 & 0.001 & \textbf{0.29518} & 0.00124 \\
recurrent---abrupt---222 & 0.32199 & 0.03015 & \textbf{0.3104} & 0.02955 \\
recurrent---abrupt---322 & 0.37013 & 0.01279 & \textbf{0.36394} & 0.02985 \\
recurrent---abrupt---332 & 0.34405 & 0.08873 & \textbf{0.32268} & 0.09904 \\
recurrent---abrupt---333 & 0.35421 & 0.00811 & \textbf{0.35155} & 0.04098 \\
recurrent---abrupt---334 & 0.39439 & 0.00079 & \textbf{0.37892} & 0.05352 \\
recurrent---abrupt---335 & 0.39358 & 0.00095 & \textbf{0.37639} & 0.04541 \\
recurrent---abrupt---422 & \textbf{0.31978} & 0.0289 & 0.32019 & 0.0455 \\
recurrent---abrupt---444 & 0.39685 & 0.00128 & \textbf{0.37445} & 0.04024 \\
recurrent---abrupt---522 & \textbf{0.33077} & 0.03754 & 0.33627 & 0.05462 \\
recurrent---abrupt---555 & 0.3999 & 0.00472 & \textbf{0.34243} & 0.01166 \\
\midrule
\textbf{Unique Wins} & \multicolumn{2}{l?}{\textbf{4}} & \multicolumn{2}{l}{\textbf{20}} \\
\midrule\multicolumn{5}{l}{A \textbf{bold} error value indicates higher accuracy, and \textit{\textbf{bold italics}} indicate a tie.}\\
\multicolumn{5}{l}{\textbf{One-tailed binomial test statistics}:  \textbf{p-value: 0.00077};   \textbf{Confidence Interval:  0.65819 --- 1} }\\

				\bottomrule
		\end{tabular}}
\end{table}

\begin{table}
	
	\caption{\label{mevsyn3} LevBag - Synthetic streams with concept drift
	} 
	\scriptsize
	\centering
	\makebox[0.8\textwidth]{
			\begin{tabular}{p{6cm}?P{1.3cm}?P{1.3cm}?P{1.3cm}?P{1.3cm}}
				\toprule
				\multicolumn{1}{c?}{\textbf{Streams}} &  \multicolumn{2}{m{2.9cm}}{\textbf{VFDT Base}} &
				\multicolumn{2}{m{2.9cm}}{\textbf{EFDT Base}}\\
				\multicolumn{1}{c?}{} & Error & Variance & Error & Variance
				\\
				\midrule
				recurrent---agrawal & 0.11408 & 0.00043 & \textbf{0.10205} & 0.00066 \\
recurrent---led & \textbf{0.26217} & 2e-04 & 0.28177 & 0.00034 \\
recurrent---randomtree & 0.06448 & 0.00103 & \textbf{0.0642} & 0.0013 \\
recurrent---sea & 0.10672 & 0.00011 & \textbf{0.10639} & 0.00011 \\
recurrent---stagger & \textbf{0.00137} & 3e-05 & 0.00141 & 3e-05 \\
recurrent---waveform & \textbf{0.15119} & 0.00016 & 0.15233 & 0.00016 \\
hyperplane---1 & \textbf{0.10714} & 0.00013 & 0.11024 & 0.00013 \\
hyperplane---2 & \textbf{0.11543} & 2e-04 & 0.11736 & 2e-04 \\
hyperplane---3 & \textbf{0.10488} & 0.00012 & 0.10881 & 0.00012 \\
hyperplane---4 & \textbf{0.10807} & 0.00031 & 0.10908 & 0.00032 \\
rbf---drift-1 & \textbf{0.0605} & 0.00015 & 0.06153 & 0.00017 \\
rbf---drift-2 & 0.08728 & 0.00029 & \textbf{0.0825} & 0.00027 \\
rbf---drift-3 & 0.08917 & 0.00037 & \textbf{0.08741} & 0.00035 \\
rbf---drift-4 & 0.18823 & 9e-04 & \textbf{0.17382} & 0.00075 \\
recurrent---abrupt---222 & \textbf{0.08568} & 0.03789 & 0.09068 & 0.03983 \\
recurrent---abrupt---322 & \textbf{0.15231} & 0.02417 & 0.15752 & 0.01998 \\
recurrent---abrupt---332 & \textbf{0.11588} & 0.04298 & 0.11655 & 0.04618 \\
recurrent---abrupt---333 & \textbf{0.15151} & 0.02658 & 0.16132 & 0.02506 \\
recurrent---abrupt---334 & 0.21071 & 0.00723 & \textbf{0.17869} & 0.03142 \\
recurrent---abrupt---335 & 0.20627 & 0.01275 & \textbf{0.1693} & 0.04913 \\
recurrent---abrupt---422 & \textbf{0.10922} & 0.04752 & 0.11094 & 0.05145 \\
recurrent---abrupt---444 & 0.17353 & 0.02142 & \textbf{0.12842} & 0.06028 \\
recurrent---abrupt---522 & \textbf{0.11904} & 0.04243 & 0.13362 & 0.05295 \\
recurrent---abrupt---555 & 0.13153 & 0.01712 & \textbf{0.08684} & 0.01581 \\
\midrule
\textbf{Unique Wins} & \multicolumn{2}{l?}{\textbf{14}} & \multicolumn{2}{l}{\textbf{10}} \\
\midrule\multicolumn{5}{l}{A \textbf{bold} error value indicates higher accuracy, and \textit{\textbf{bold italics}} indicate a tie.}\\
\multicolumn{5}{l}{\textbf{One-tailed binomial test statistics}:  \textbf{p-value: 0.84627};   \textbf{Confidence Interval:  0.24639 --- 1} }\\

				\bottomrule
		\end{tabular}}
	
\end{table}

\begin{table}[!ht]
	
	\caption{\label{mevsyn4} Adaptive Random Forest - Synthetic streams with concept drift
	} 
	\scriptsize
	\centering
	\makebox[0.8\textwidth]{
			\begin{tabular}{p{6cm}?P{1.3cm}?P{1.3cm}?P{1.3cm}?P{1.3cm}}
				\toprule
				\multicolumn{1}{c?}{\textbf{Streams}} &  \multicolumn{2}{m{2.6cm}}{\textbf{VFDT Base}} &
				\multicolumn{2}{m{2.6cm}}{\textbf{EFDT Base}}\\
				\multicolumn{1}{c?}{} & Error & Variance & Error & Variance 
				\\
				\midrule
				recurrent---agrawal & 0.34164 & 0.00079 & \textbf{0.23287} & 0.00131 \\
recurrent---led & 0.28165 & 0.00065 & \textbf{0.26301} & 2e-04 \\
recurrent---randomtree & 0.24865 & 0.0037 & \textbf{0.21096} & 0.00363 \\
recurrent---sea & 0.15533 & 0.00024 & \textbf{0.14944} & 0.00023 \\
recurrent---stagger & 0.08174 & 0.00094 & \textbf{0.00365} & 0.00028 \\
recurrent---waveform & \textbf{0.15365} & 0.00017 & 0.15474 & 0.00016 \\
hyperplane---1 & 0.13866 & 3e-04 & \textbf{0.13727} & 0.00015 \\
hyperplane---2 & 0.13581 & 3e-04 & \textbf{0.13454} & 0.00026 \\
hyperplane---3 & 0.14088 & 0.00021 & \textbf{0.13967} & 0.00015 \\
hyperplane---4 & 0.14896 & 0.00222 & \textbf{0.12455} & 0.00042 \\
rbf---drift-1 & 0.17788 & 0.00129 & \textbf{0.16278} & 0.00143 \\
rbf---drift-2 & 0.16438 & 0.00119 & \textbf{0.14905} & 0.00121 \\
rbf---drift-3 & 0.16781 & 0.0011 & \textbf{0.15322} & 0.00115 \\
rbf---drift-4 & 0.20026 & 0.00079 & \textbf{0.18832} & 0.00076 \\
recurrent---abrupt---222 & 0.11886 & 0.01174 & \textbf{0.00548} & 0.00033 \\
recurrent---abrupt---322 & 0.07783 & 0.00916 & \textbf{0.0182} & 0.00182 \\
recurrent---abrupt---332 & 0.16737 & 0.01644 & \textbf{0.06472} & 0.01184 \\
recurrent---abrupt---333 & 0.21102 & 0.01086 & \textbf{0.11341} & 0.0059 \\
recurrent---abrupt---334 & 0.22163 & 0.01839 & \textbf{0.15657} & 0.00494 \\
recurrent---abrupt---335 & 0.28508 & 0.01309 & \textbf{0.15258} & 0.00231 \\
recurrent---abrupt---422 & 0.09237 & 0.01104 & \textbf{0.01034} & 0.00065 \\
recurrent---abrupt---444 & 0.43723 & 0.01824 & \textbf{0.21796} & 0.01087 \\
recurrent---abrupt---522 & 0.06868 & 0.00863 & \textbf{0.01917} & 0.00181 \\
recurrent---abrupt---555 & 0.67892 & 0.00197 & \textbf{0.58899} & 0.01103 \\
\midrule
\textbf{Unique Wins} & \multicolumn{2}{l?}{\textbf{1}} & \multicolumn{2}{l}{\textbf{23}} \\
\midrule\multicolumn{5}{l}{A \textbf{bold} error value indicates higher accuracy, and \textit{\textbf{bold italics}} indicate a tie.}\\
\multicolumn{5}{l}{\textbf{One-tailed binomial test statistics}:  \textbf{p-value: $<$ 0.00001};   \textbf{Confidence Interval:  0.81711 --- 1} }\\

				\bottomrule
		\end{tabular}}
\end{table}

	With bagging ensembles, EFDT as a base learner demonstrates an advantage with OzaBag (Table \ref{mevsyn0}, p-value 0.00331),  Leveraging Bagging without ADWIN (Table \ref{mevsyn2}, p-value 0.00077), and Adaptive Random Forest (Table \ref{mevsyn4}, p-value $<$ 0.00001). EFDT has no advantage as a base learner with OzaBagADWIN (Table \ref{mevsyn1}, p-value 0.41941, arising from 13 wins for EFDT to 11 for VFDT) and Leveraging Bagging (Table \ref{mevsyn3}, p-value 0.84627 arising from 10 wins for EFDT to 14 for VFDT). 
	
	These two cases may be explained as follows. When EFDT replaces a test, there is a risk that the model as a whole will derease in accuracy in the short term, as the substrees that have been removed, while suboptimal, may be better than the simple split with which they are replaced. Ensembles with change detection use ADWIN change detectors to determine if change is occurring, and if so, replace trees with poor performance. The change detectors are triggered when accuracy drops. When there is drift, EFDT is likely to replace nodes (and thus corresponding subtrees)---triggering change detectors due to loss of accuracy. Thus, EFDT's response to drift, in which it is already adjusting the tree to the new distribution, will trigger total removal of the tree.
	
	Therefore in settings with concept drift (as all our synthetic streams are), where ensembles happen to feature both a bagging component and an ADWIN change detector, a tree created by EFDT is more likely to be removed when a change is detected without the tree growing large enough or persisting long enough to meaningfully contribute to prediction.
	
	
\begin{table}
	
	\caption{\label{mevsyn5} OzaBoost - Synthetic streams with concept drift
	} 
	\scriptsize
	\centering
	\makebox[0.8\textwidth]{
			\begin{tabular}{p{6cm}?P{1.3cm}?P{1.3cm}?P{1.3cm}?P{1.3cm}}
				\toprule
				\multicolumn{1}{c?}{\textbf{Streams}} &  \multicolumn{2}{m{2.9cm}}{\textbf{VFDT Base}} & 
				\multicolumn{2}{m{2.9cm}}{\textbf{EFDT Base}}\\
				\multicolumn{1}{c?}{} & Error & Variance & Error & Variance \\
				\\
				\midrule
				recurrent---agrawal & 0.1812 & 0.00087 & \textbf{0.15742} & 0.00042 \\
recurrent---led & \textbf{0.33442} & 0.00043 & 0.33533 & 0.00041 \\
recurrent---randomtree & 0.14317 & 0.00089 & \textbf{0.13294} & 0.00115 \\
recurrent---sea & 0.13049 & 0.00012 & \textbf{0.12719} & 0.00013 \\
recurrent---stagger & 0.10382 & 0.00238 & \textbf{0.08442} & 0.00194 \\
recurrent---waveform & 0.17132 & 0.00086 & \textbf{0.16516} & 0.00044 \\
hyperplane---1 & \textbf{0.10869} & 0.00014 & 0.11215 & 0.00016 \\
hyperplane---2 & 0.13259 & 0.00037 & \textbf{0.12523} & 0.00028 \\
hyperplane---3 & \textbf{0.10511} & 0.00012 & 0.11043 & 0.00015 \\
hyperplane---4 & 0.12736 & 0.00093 & \textbf{0.11864} & 0.00058 \\
rbf---drift-1 & 0.06847 & 0.00018 & \textbf{0.06664} & 0.00017 \\
rbf---drift-2 & 0.14149 & 0.00039 & \textbf{0.13559} & 0.00036 \\
rbf---drift-3 & 0.10352 & 0.00039 & \textbf{0.10126} & 0.00038 \\
rbf---drift-4 & 0.31758 & 0.00094 & \textbf{0.31195} & 0.00088 \\
recurrent---abrupt---222 & 0.24091 & 0.08901 & \textbf{0.22166} & 0.08186 \\
recurrent---abrupt---322 & 0.21457 & 0.05245 & \textbf{0.17571} & 0.06421 \\
recurrent---abrupt---332 & 0.2061 & 0.05159 & \textbf{0.09681} & 0.04476 \\
recurrent---abrupt---333 & 0.21236 & 0.01179 & \textbf{0.11988} & 0.02705 \\
recurrent---abrupt---334 & 0.21905 & 0.00198 & \textbf{0.10993} & 0.02107 \\
recurrent---abrupt---335 & 0.21731 & 0.00135 & \textbf{0.12118} & 0.02446 \\
recurrent---abrupt---422 & 0.2311 & 0.08934 & \textbf{0.18369} & 0.08844 \\
recurrent---abrupt---444 & 0.2074 & 0.013 & \textbf{0.05954} & 0.00795 \\
recurrent---abrupt---522 & 0.2181 & 0.06227 & \textbf{0.18467} & 0.06745 \\
recurrent---abrupt---555 & 0.3756 & 0.02529 & \textbf{0.11939} & 0.00508 \\
\midrule
\textbf{Unique Wins} & \multicolumn{2}{l?}{\textbf{3}} & \multicolumn{2}{l}{\textbf{21}} \\
\midrule\multicolumn{5}{l}{A \textbf{bold} error value indicates higher accuracy, and \textit{\textbf{bold italics}} indicate a tie.}\\
\multicolumn{5}{l}{\textbf{One-tailed binomial test statistics}:  \textbf{p-value: 0.00014};   \textbf{Confidence Interval:  0.70773 --- 1} }\\

				\bottomrule
		\end{tabular}}
\end{table}

\begin{table}
	
	\caption{\label{mevsyn6} OzaBoostADWIN - Synthetic streams with concept drift
	} 
	\scriptsize
	\centering
	\makebox[0.8\textwidth]{
			\begin{tabular}{p{6cm}?P{1.3cm}?P{1.3cm}?P{1.3cm}?P{1.3cm}}
				\toprule
				\multicolumn{1}{c?}{\textbf{Streams}} &  \multicolumn{2}{m{2.9cm}}{\textbf{VFDT Base}} &
				\multicolumn{2}{m{2.9cm}}{\textbf{EFDT Base}}\\
				\multicolumn{1}{c?}{} & Error & Variance & Error & Variance
				\\
				\midrule
				recurrent---agrawal & \textbf{0.16874} & 0.001 & 0.1748 & 0.00088 \\
recurrent---led & \textbf{0.27943} & 6e-04 & 0.28262 & 0.00156 \\
recurrent---randomtree & 0.13345 & 0.00305 & \textbf{0.12445} & 0.00323 \\
recurrent---sea & 0.16147 & 0.00044 & \textbf{0.1449} & 0.00038 \\
recurrent---stagger & \textbf{0.00152} & 0.00011 & 0.01347 & 0.00807 \\
recurrent---waveform & 0.18972 & 0.00032 & \textbf{0.18629} & 0.00031 \\
hyperplane---1 & 0.16723 & 0.00032 & \textbf{0.15898} & 0.00028 \\
hyperplane---2 & 0.188 & 0.00051 & \textbf{0.16057} & 0.00042 \\
hyperplane---3 & 0.16682 & 3e-04 & \textbf{0.15808} & 0.00033 \\
hyperplane---4 & 0.17259 & 0.00081 & \textbf{0.14838} & 0.00062 \\
rbf---drift-1 & 0.07998 & 0.00027 & \textbf{0.07398} & 0.00022 \\
rbf---drift-2 & 0.13404 & 0.00096 & \textbf{0.11857} & 0.00057 \\
rbf---drift-3 & 0.11905 & 0.00052 & \textbf{0.10793} & 0.00039 \\
rbf---drift-4 & 0.2341 & 0.00145 & \textbf{0.22421} & 0.00128 \\
recurrent---abrupt---222 & \textbf{0.07473} & 0.0385 & 0.09672 & 0.04831 \\
recurrent---abrupt---322 & \textbf{0.01106} & 0.00578 & 0.03094 & 0.02574 \\
recurrent---abrupt---332 & 0.00034 & 4e-05 & \textbf{0.00031} & 4e-05 \\
recurrent---abrupt---333 & 0.00053 & 1e-05 & \textbf{0.00042} & 0 \\
recurrent---abrupt---334 & 0.00243 & 0.00014 & \textbf{0.00093} & 2e-05 \\
recurrent---abrupt---335 & 0.00475 & 0.00113 & \textbf{0.0023} & 3e-04 \\
recurrent---abrupt---422 & 0.08024 & 0.0799 & \textbf{0.02032} & 0.02003 \\
recurrent---abrupt---444 & 0.04236 & 0.01526 & \textbf{0.01399} & 0.00533 \\
recurrent---abrupt---522 & \textbf{0.01728} & 0.01396 & 0.0188 & 0.01558 \\
recurrent---abrupt---555 & 0.41908 & 0.06705 & \textbf{0.17661} & 0.02764 \\
\midrule
\textbf{Unique Wins} & \multicolumn{2}{l?}{\textbf{6}} & \multicolumn{2}{l}{\textbf{18}} \\
\midrule\multicolumn{5}{l}{A \textbf{bold} error value indicates higher accuracy, and \textit{\textbf{bold italics}} indicate a tie.}\\
\multicolumn{5}{l}{\textbf{One-tailed binomial test statistics}:  \textbf{p-value: 0.01133};   \textbf{Confidence Interval:  0.56531 --- 1} }\\

				\bottomrule
		\end{tabular}}
\end{table}

\begin{table}
	
	\caption{\label{mevsyn7} ADOB - Synthetic streams with concept drift
	} 
	\scriptsize
	\centering
	\makebox[0.8\textwidth]{
			\begin{tabular}{p{6cm}?P{1.3cm}?P{1.3cm}?P{1.3cm}?P{1.3cm}}
				\toprule
				\multicolumn{1}{c?}{\textbf{Streams}} &  \multicolumn{2}{m{2.9cm}}{\textbf{VFDT Base}} &
				\multicolumn{2}{m{2.9cm}}{\textbf{EFDT Base}}\\
				\multicolumn{1}{c?}{} & Error & Variance & Error & Variance
				\\
				\midrule
				recurrent---agrawal & 0.17084 & 0.00025 & \textbf{0.15346} & 0.00019 \\
recurrent---led & 0.31423 & 0.00025 & \textbf{0.30981} & 0.00041 \\
recurrent---randomtree & 0.13912 & 0.00104 & \textbf{0.1323} & 0.00104 \\
recurrent---sea & 0.12015 & 0.00011 & \textbf{0.11842} & 0.00011 \\
recurrent---stagger & 0.09834 & 0.00252 & \textbf{0.06311} & 0.00206 \\
recurrent---waveform & 0.17092 & 0.00075 & \textbf{0.16448} & 0.00043 \\
hyperplane---1 & \textbf{0.11168} & 0.00012 & 0.11396 & 0.00013 \\
hyperplane---2 & 0.12744 & 0.00027 & \textbf{0.12698} & 0.00025 \\
hyperplane---3 & \textbf{0.11004} & 0.00013 & 0.11241 & 0.00012 \\
hyperplane---4 & 0.12168 & 0.00057 & \textbf{0.11813} & 0.00046 \\
rbf---drift-1 & 0.06694 & 0.00017 & \textbf{0.06479} & 0.00016 \\
rbf---drift-2 & 0.13939 & 0.00039 & \textbf{0.13468} & 0.00037 \\
rbf---drift-3 & 0.10173 & 0.00036 & \textbf{0.09913} & 0.00034 \\
rbf---drift-4 & 0.3171 & 0.00092 & \textbf{0.31098} & 0.00089 \\
recurrent---abrupt---222 & \textbf{0.12251} & 0.03055 & 0.12888 & 0.03745 \\
recurrent---abrupt---322 & 0.15083 & 0.05514 & \textbf{0.12394} & 0.04509 \\
recurrent---abrupt---332 & 0.12366 & 0.02724 & \textbf{0.05914} & 0.02407 \\
recurrent---abrupt---333 & 0.17282 & 0.01709 & \textbf{0.10713} & 0.01841 \\
recurrent---abrupt---334 & 0.17084 & 0.00555 & \textbf{0.08981} & 0.01511 \\
recurrent---abrupt---335 & 0.1672 & 0.00628 & \textbf{0.09548} & 0.01524 \\
recurrent---abrupt---422 & 0.15191 & 0.04498 & \textbf{0.14134} & 0.04526 \\
recurrent---abrupt---444 & 0.11911 & 0.00339 & \textbf{0.04105} & 0.00638 \\
recurrent---abrupt---522 & \textbf{0.1249} & 0.03142 & 0.13476 & 0.04532 \\
recurrent---abrupt---555 & 0.32364 & 0.0243 & \textbf{0.08112} & 0.00314 \\
\midrule
\textbf{Unique Wins} & \multicolumn{2}{l?}{\textbf{4}} & \multicolumn{2}{l}{\textbf{20}} \\
\midrule\multicolumn{5}{l}{A \textbf{bold} error value indicates higher accuracy, and \textit{\textbf{bold italics}} indicate a tie.}\\
\multicolumn{5}{l}{\textbf{One-tailed binomial test statistics}:  \textbf{p-value: 0.00077};   \textbf{Confidence Interval:  0.65819 --- 1} }\\

				\bottomrule
		\end{tabular}}
\end{table}

\begin{table}
	
	\caption{\label{mevsyn8} BOLE - Synthetic streams with concept drift
	} 
	\scriptsize
	\centering
	\makebox[0.8\textwidth]{
			\begin{tabular}{p{6cm}?P{1.3cm}?P{1.3cm}?P{1.3cm}?P{1.3cm}}
				\toprule
				\multicolumn{1}{c?}{\textbf{Streams}} &  \multicolumn{2}{m{2.9cm}}{\textbf{VFDT Base}} &
				\multicolumn{2}{m{2.9cm}}{\textbf{EFDT Base}}\\
				\multicolumn{1}{c?}{} & Error & Variance & Error & Variance
				\\
				\midrule
				recurrent---agrawal & 0.17084 & 0.00025 & \textbf{0.15346} & 0.00019 \\
recurrent---led & 0.31417 & 0.00025 & \textbf{0.30975} & 0.00041 \\
recurrent---randomtree & 0.13912 & 0.00104 & \textbf{0.1323} & 0.00104 \\
recurrent---sea & 0.12015 & 0.00011 & \textbf{0.11841} & 0.00011 \\
recurrent---stagger & 0.09834 & 0.00252 & \textbf{0.06311} & 0.00206 \\
recurrent---waveform & 0.17092 & 0.00075 & \textbf{0.16447} & 0.00043 \\
hyperplane---1 & \textbf{0.11168} & 0.00012 & 0.11396 & 0.00013 \\
hyperplane---2 & 0.12744 & 0.00027 & \textbf{0.12698} & 0.00025 \\
hyperplane---3 & \textbf{0.11004} & 0.00013 & 0.11241 & 0.00012 \\
hyperplane---4 & 0.12168 & 0.00057 & \textbf{0.11813} & 0.00046 \\
rbf---drift-1 & 0.06694 & 0.00017 & \textbf{0.06479} & 0.00016 \\
rbf---drift-2 & 0.13939 & 0.00039 & \textbf{0.13468} & 0.00037 \\
rbf---drift-3 & 0.10173 & 0.00036 & \textbf{0.09913} & 0.00034 \\
rbf---drift-4 & 0.3171 & 0.00092 & \textbf{0.31099} & 0.00089 \\
recurrent---abrupt---222 & \textbf{0.12248} & 0.03055 & 0.12743 & 0.03735 \\
recurrent---abrupt---322 & 0.1498 & 0.05508 & \textbf{0.12291} & 0.04503 \\
recurrent---abrupt---332 & 0.12366 & 0.02724 & \textbf{0.05914} & 0.02407 \\
recurrent---abrupt---333 & 0.17282 & 0.01709 & \textbf{0.10713} & 0.01841 \\
recurrent---abrupt---334 & 0.17083 & 0.00555 & \textbf{0.08965} & 0.01502 \\
recurrent---abrupt---335 & 0.16706 & 0.00622 & \textbf{0.0949} & 0.01492 \\
recurrent---abrupt---422 & 0.15191 & 0.04498 & \textbf{0.14134} & 0.04526 \\
recurrent---abrupt---444 & 0.11737 & 0.00302 & \textbf{0.03993} & 0.00592 \\
recurrent---abrupt---522 & \textbf{0.12386} & 0.03136 & 0.13373 & 0.04526 \\
recurrent---abrupt---555 & 0.29368 & 0.01059 & \textbf{0.07874} & 0.00196 \\
\midrule
\textbf{Unique Wins} & \multicolumn{2}{l?}{\textbf{4}} & \multicolumn{2}{l}{\textbf{20}} \\
\midrule\multicolumn{5}{l}{A \textbf{bold} error value indicates higher accuracy, and \textit{\textbf{bold italics}} indicate a tie.}\\
\multicolumn{5}{l}{\textbf{One-tailed binomial test statistics}:  \textbf{p-value: 0.00077};   \textbf{Confidence Interval:  0.65819 --- 1} }\\

				\bottomrule
		\end{tabular}}
\end{table}

\begin{table}
	
	\caption{\label{mevsyn9} OnlineSmoothBoost - Synthetic streams with concept drift
	} 
	\scriptsize
	\centering
	\makebox[0.8\textwidth]{
		\begin{tabular}{p{6cm}?P{1.3cm}?P{1.3cm}?P{1.3cm}?P{1.3cm}}
			\toprule
			\multicolumn{1}{c?}{\textbf{Streams}} &  \multicolumn{2}{m{2.9cm}}{\textbf{VFDT Base}} &
			\multicolumn{2}{m{2.9cm}}{\textbf{EFDT Base}}\\
				\multicolumn{1}{c?}{} & Error & Variance & Error & Variance
				\\
			\midrule
			recurrent---agrawal & 0.19145 & 0.00037 & \textbf{0.17937} & 0.00038 \\
recurrent---led & \textbf{0.31266} & 0.00027 & 0.32583 & 0.00036 \\
recurrent---randomtree & 0.20745 & 0.00268 & \textbf{0.19807} & 0.0023 \\
recurrent---sea & 0.1442 & 0.00013 & \textbf{0.13578} & 0.00013 \\
recurrent---stagger & \textbf{0.17514} & 0.00101 & 0.18746 & 0.00046 \\
recurrent---waveform & 0.1712 & 0.00148 & \textbf{0.16125} & 0.00119 \\
hyperplane---1 & 0.10209 & 0.00017 & \textbf{0.098} & 0.00013 \\
hyperplane---2 & 0.14834 & 0.00104 & \textbf{0.12153} & 0.00054 \\
hyperplane---3 & 0.09573 & 0.00011 & \textbf{0.09475} & 1e-04 \\
hyperplane---4 & 0.14618 & 0.00297 & \textbf{0.1171} & 0.00112 \\
rbf---drift-1 & 0.084 & 0.00035 & \textbf{0.07305} & 0.00027 \\
rbf---drift-2 & 0.18332 & 0.00084 & \textbf{0.15977} & 0.00076 \\
rbf---drift-3 & 0.11502 & 0.00058 & \textbf{0.10907} & 0.00057 \\
rbf---drift-4 & 0.33978 & 0.00122 & \textbf{0.33032} & 0.0012 \\
recurrent---abrupt---222 & 0.33921 & 0.02677 & \textbf{0.33905} & 0.02662 \\
recurrent---abrupt---322 & \textbf{0.3587} & 0.01792 & 0.36061 & 0.02554 \\
recurrent---abrupt---332 & 0.32643 & 0.05302 & \textbf{0.30822} & 0.06033 \\
recurrent---abrupt---333 & 0.35755 & 0.0149 & \textbf{0.34818} & 0.0266 \\
recurrent---abrupt---334 & 0.38918 & 0.00202 & \textbf{0.35711} & 0.03683 \\
recurrent---abrupt---335 & 0.38905 & 0.00225 & \textbf{0.37848} & 0.02378 \\
recurrent---abrupt---422 & 0.32939 & 0.03697 & \textbf{0.32422} & 0.04199 \\
recurrent---abrupt---444 & 0.38303 & 0.00529 & \textbf{0.36083} & 0.01028 \\
recurrent---abrupt---522 & \textbf{0.32769} & 0.04879 & 0.32929 & 0.04613 \\
recurrent---abrupt---555 & 0.39253 & 0.00499 & \textbf{0.34937} & 0.0076 \\
\midrule
\textbf{Unique Wins} & \multicolumn{2}{l?}{\textbf{4}} & \multicolumn{2}{l}{\textbf{20}} \\
\midrule\multicolumn{5}{l}{A \textbf{bold} error value indicates higher accuracy, and \textit{\textbf{bold italics}} indicate a tie.}\\
\multicolumn{5}{l}{\textbf{One-tailed binomial test statistics}:  \textbf{p-value: 0.00077};   \textbf{Confidence Interval:  0.65819 --- 1} }\\

			\bottomrule
		\end{tabular}}
\end{table}

\begin{table}
	
	\caption{\label{mevsyn10} Plain single learners (no ensemble) - Synthetic streams with concept drift
	} 
	\scriptsize
	\centering
	\makebox[0.8\textwidth]{
		\begin{tabular}{p{6cm}?P{1.3cm}?P{1.3cm}?P{1.3cm}?P{1.3cm}}
			\toprule
			\multicolumn{1}{c?}{\textbf{Streams}} &  \multicolumn{2}{m{2.9cm}}{\textbf{VFDT Base}} &
			\multicolumn{2}{m{2.9cm}}{\textbf{EFDT Base}}\\
				\multicolumn{1}{c?}{} & Error & Variance & Error & Variance
				\\
			\midrule
			recurrent---agrawal & 0.20846 & 0.00043 & \textbf{0.19639} & 0.00062 \\
recurrent---led & \textbf{0.33838} & 0.00068 & 0.34679 & 0.00037 \\
recurrent---randomtree & 0.22404 & 0.00231 & \textbf{0.21817} & 0.00219 \\
recurrent---sea & 0.15251 & 0.00016 & \textbf{0.14958} & 0.00019 \\
recurrent---stagger & \textbf{0.1882} & 0.00047 & 0.19043 & 0.00047 \\
recurrent---waveform & 0.19355 & 0.00171 & \textbf{0.18957} & 0.00135 \\
hyperplane---1 & \textbf{0.11566} & 0.00021 & 0.11677 & 2e-04 \\
hyperplane---2 & 0.16785 & 0.00117 & \textbf{0.14056} & 0.00069 \\
hyperplane---3 & \textbf{0.1074} & 0.00013 & 0.11208 & 0.00015 \\
hyperplane---4 & 0.16309 & 0.00359 & \textbf{0.13384} & 0.00147 \\
rbf---drift-1 & 0.11462 & 0.00053 & \textbf{0.11255} & 0.00063 \\
rbf---drift-2 & 0.2858 & 0.00155 & \textbf{0.2623} & 0.00171 \\
rbf---drift-3 & \textbf{0.13821} & 0.00068 & 0.14308 & 0.00087 \\
rbf---drift-4 & 0.40874 & 0.00141 & \textbf{0.40597} & 0.00144 \\
recurrent---abrupt---222 & 0.35403 & 0.02056 & \textbf{0.35381} & 0.02053 \\
recurrent---abrupt---322 & 0.37862 & 0.01248 & \textbf{0.37596} & 0.03198 \\
recurrent---abrupt---332 & 0.3504 & 0.08913 & \textbf{0.33376} & 0.10312 \\
recurrent---abrupt---333 & \textbf{0.36505} & 0.01499 & 0.36583 & 0.04067 \\
recurrent---abrupt---334 & 0.39687 & 0.00217 & \textbf{0.39001} & 0.04158 \\
recurrent---abrupt---335 & 0.39622 & 0.00319 & \textbf{0.3945} & 0.02991 \\
recurrent---abrupt---422 & \textbf{0.33416} & 0.02931 & 0.33569 & 0.04239 \\
recurrent---abrupt---444 & 0.40671 & 0.00636 & \textbf{0.38967} & 0.02007 \\
recurrent---abrupt---522 & \textbf{0.3309} & 0.03999 & 0.33374 & 0.04749 \\
recurrent---abrupt---555 & 0.46461 & 0.00584 & \textbf{0.40866} & 0.01097 \\
\midrule
\textbf{Unique Wins} & \multicolumn{2}{l?}{\textbf{8}} & \multicolumn{2}{l}{\textbf{16}} \\
\midrule\multicolumn{5}{l}{A \textbf{bold} error value indicates higher accuracy, and \textit{\textbf{bold italics}} indicate a tie.}\\
\multicolumn{5}{l}{\textbf{One-tailed binomial test statistics}:  \textbf{p-value: 0.07579};   \textbf{Confidence Interval:  0.47858 --- 1} }\\

			\bottomrule
		\end{tabular}}
\end{table}

	EFDT based ensembles demonstrate superior prequential accuracy performance within a significance level of 0.05 when used as a base learner with all boosting strategies: OzaBoost (Table \ref{mevsyn5}, p-value 0.00014), OzaBoostADWIN (Table \ref{mevsyn6}, p-value 0.01133), ADOB, BOLE and OnlineSmoothBoost (Tables \ref{mevsyn7}, \ref{mevsyn8}, and \ref{mevsyn9}, all three p-values 0.00077 on account of 20 wins and 4 losses).
	
	Plain EFDT outperforms plain VFDT on 16 synthetic streams (Table \ref{mevsyn10}, with VFDT outperforming on the remaining 8.
	
	The ranked p-values are $<$ 0.00001, 0.00014, 0.00077, 0.00077, 0.00077, 0.00077, 0.00331, 0.01133, 0.41941, 0.84627. The first eight null hypotheses would be rejected using a Holm-Bonferroni multiple testing correction, resulting in findings of significant improvements in the number of synthetic non-stationary data streams for which the ensemble technique is more accurate using EFDT than VFDT for OzaBag, LevBagAdwin, Adaptive Random Forest, OzaBoost, OzaBoostADWIN, ADOB, BOLE and OnlineSmoothBoost.
	
	\FloatBarrier

	\section{Conclusions}
	\label{conclusions}

	\begin{table}
		
		\caption{\label{summary} Summary of results
		} 
		\scriptsize
		\centering
		\makebox[0.8\textwidth]{
			\scalebox{0.85}{
				\begin{tabular}{p{3cm}?P{1cm}?P{1cm}?P{1.3cm}?P{1cm}?P{1cm}?P{1.3cm}?P{1cm}?P{1cm}?P{1.3cm}}
					\toprule
					\multicolumn{1}{c?}{\textbf{Ensembles}} &  \multicolumn{3}{m{3.3cm}}{\textbf{UCI Streams }} & \multicolumn{3}{m{3.3cm}}{\textbf{UCI Shuffled Streams}} & \multicolumn{3}{m{3.3cm}}{\textbf{Synthetic Streams}} \\
					& VFDT wins & EFDT wins & p-value & VFDT wins & EFDT wins & p-value & VFDT wins & EFDT wins & p-value
					\\
					\midrule
					OzaBag & 2 & 16 & 0.00066 & 3 & 15 & 0.00377 & 5 & 19 & 0.00331 \\
OzaBagAdwin & 3 & 15 & 0.00377 & 3 & 15 & 0.00377 & 11 & 13 & 0.41941 \\
LevBag without Adwin & 6 & 13 & 0.08353 & 7 & 11 & 0.24034 & 4 & 20 & 0.00077 \\
LevBag & 5 & 13 & 0.04813 & 10 & 8 & 0.75966 & 14 & 10 & 0.84627 \\
ARF & 3 & 15 & 0.00377 & 4 & 16 & 0.00591 & 1 & 23 & 0 \\
OzaBoost & 6 & 12 & 0.11894 & 7 & 10 & 0.31453 & 3 & 21 & 0.00014 \\
OzaBoostAdwin & 8 & 11 & 0.3238 & 7 & 11 & 0.24034 & 6 & 18 & 0.01133 \\
ADOB & 5 & 13 & 0.04813 & 5 & 13 & 0.04813 & 4 & 20 & 0.00077 \\
BOLE & 5 & 13 & 0.04813 & 5 & 13 & 0.04813 & 4 & 20 & 0.00077 \\
OnlineSmoothBoost & 5 & 15 & 0.02069 & 3 & 15 & 0.00377 & 4 & 20 & 0.00077 \\
Plain (no ensemble) & 7 & 13 & 0.13159 & 5 & 13 & 0.04813 & 8 & 16 & 0.07579 \\

					\midrule	
					
					\multicolumn{9}{p{15cm}}{EFDT largely outperforms VFDT as a base learner for ensembles, achieving significance at a standard 0.05 level with \textbf{7 ensembles} on UCI streams, \textbf{6 ensembles} on shuffled UCI streams, and \textbf{8 ensembles} on the synthetic testbench.} & \multicolumn{1}{m{1cm}}{} 
					
					\\ 
					\multicolumn{9}{p{15cm}}{} & \multicolumn{1}{m{1cm}}{}
					
					\\
					
					\multicolumn{9}{p{15cm}}{The test is a one-tailed binomial test to determine the probability that EFDT-based ensembles would achieve so many wins if wins and losses were equiprobable.} & \multicolumn{1}{m{1cm}}{} \\
					
					\bottomrule
			\end{tabular}}
		}
	\end{table}

	Table \ref{summary} summarises our results. EFDT has an advantage relative to VFDT in 28 out of 30 ensemble/stream combination settings when employed as the base learner for online ensembling techniques. For 21 of these settings, the advantage is within a statistically significant level of 0.05. For the 2 (out of 30) ensemble/scenario combinations for which EFDT ``loses'' (LevBag with synthetic and shuffled UCI streams), the win for VFDT does not reach statistical significance.  If a multiple testing correction is applied within each family of tests, EFDT as a base learner achieves higher accuracy significantly more often than VFDT for 14 out of the 30 ensemble/scenario combinations.  Both with and without multiple testing corrections, VFDT is never more accurate significantly more often than EFDT.
	
	EFDT shows promise as a base learner for both boosting and bagging ensembles in synthetic concept-drifting scenarios, achieving lower error than VFDT within a $0.05$ significance level with 8 out of 10 ensembles (7 if a multiple testing correction is applied). However, in concept-drifting scenarios, given EFDT's greater short term instability, interactions between EFDT and change detectors may cause some EFDT ensemble components to be prematurely terminated, leading to an erosion of advantage. This interaction needs further study.
	The influence of change detectors appears to be negligible when concept drift is not present.
	
	On UCI data streams that are not shuffled, using EFDT as a base learner leads to significant wins for 7 ensembles at the $0.05$ significance level, with the remaining results falling outside the level of significance. Results on shuffled versions of the streams are similar, with significant wins for EFDT on 6 ensembles. Bagging ensembles are overwhelmingly favored, with four out of five achieving significance for unshuffled streams; however, three boosting strategies, ADOB, BOLE, and OnlineSmoothBoost, with rationalized weighting mechanisms also lead to a win for the EFDT base learner within the $0.05$ significance level. The indication seems to be that rationalized regimes for weighting, as noted in ADOB, BOLE and OnlineSmoothBoost, interact positively with HATT to deliver a performance gain over the boosting regimes that weight more naively---OzaBoost and OzaBoostAdwin.
	
	The observations that EFDT does not work well with change detectors that are based on model error under concept drift, and that EFDT can revise models without requiring a change detector both suggest that bespoke EFDT-based ensemble methods might be effective under concept drift. Existing online ensemble techniques have been developed for the rigid unrevisable models of VFDT. New online ensemble techniques that exploit the flexibility of EFDT provide a promising direction for future research.
	
	Tangentially, while we have not explicitly compared the general performance of bagging and boosting strategies in this paper, as our focus is on the impact of using EFDT as a base learner in ensemble techniques, we offer the hypothesis that strategies that rationalize weighting in boosting meta-algorithms using base learners with limited instability might reduce the performance gap between online bagging and online boosting ensembles pointed out in \cite{bifet2010leveraging}.
	
	Leveraging Bagging is significantly advantaged by an EFDT base learner over a VFDT base learner when mild concept drift is present, as in UCI streams, but the advantage erodes when one moves to a scenario with a more uniform, non-evolving stream, as noted in Section \ref{sec:shuffled}. This erosion appears to result mostly from the change of Poisson $\lambda$ value to 6 (from 1, thus ensuring 34\% of instances are not discarded by drawing a 0), hinting at the thesis that when the generating distribution is uniform, online bagging with EFDT is advantaged through a batch-like bagging process that mimics leaving out a third of the input instances for each base learner---but when mild concept drift is present the extra instances are more helpful for learning. In a scenario with larger and more continual concept drift, as with our synthetic streams, interaction with change detectors appears to work adversely for EFDT with Leveraging Bagging.
	
	We posit that the outperformance of ensembles with HATT as a base learner over those with HT as a base learner is due to the greater amount of short term instability in HATT, allowing for greater component diversity---which is associated with more effective error reduction---making HATT particularly suitable for use as a base learner in ensembled approaches.
	
	To sum up, our results show that Hoeffding AnyTime Tree (implemented as Extremely Fast Decision Tree, EFDT) significantly outperforms Hoeffding Tree (implemented as Very Fast Decision Tree, VFDT) on prequential accuracy as a base learner for bagging and boosting ensembles on a large set of real and synthetic testbenches, and never underperforms with significance. 
	
	\printbibliography
	

\end{document}